\documentclass[11pt]{article}

\usepackage[preprint]{acl}

\usepackage{times}
\usepackage{amsmath}
\usepackage{latexsym}

\usepackage[T1]{fontenc}

\usepackage[utf8]{inputenc}

\usepackage{microtype}

\usepackage{inconsolata}

\usepackage{graphicx}

\usepackage{array}
\usepackage{booktabs}
\usepackage{multirow}
\usepackage{colortbl}  
\usepackage{subcaption}
\usepackage{placeins}
\usepackage{float}

\usepackage{graphicx}
\usepackage{hyperref}

\newcommand{\hficon}{%
  \raisebox{-0.1\height}{\includegraphics[height=1.2em]{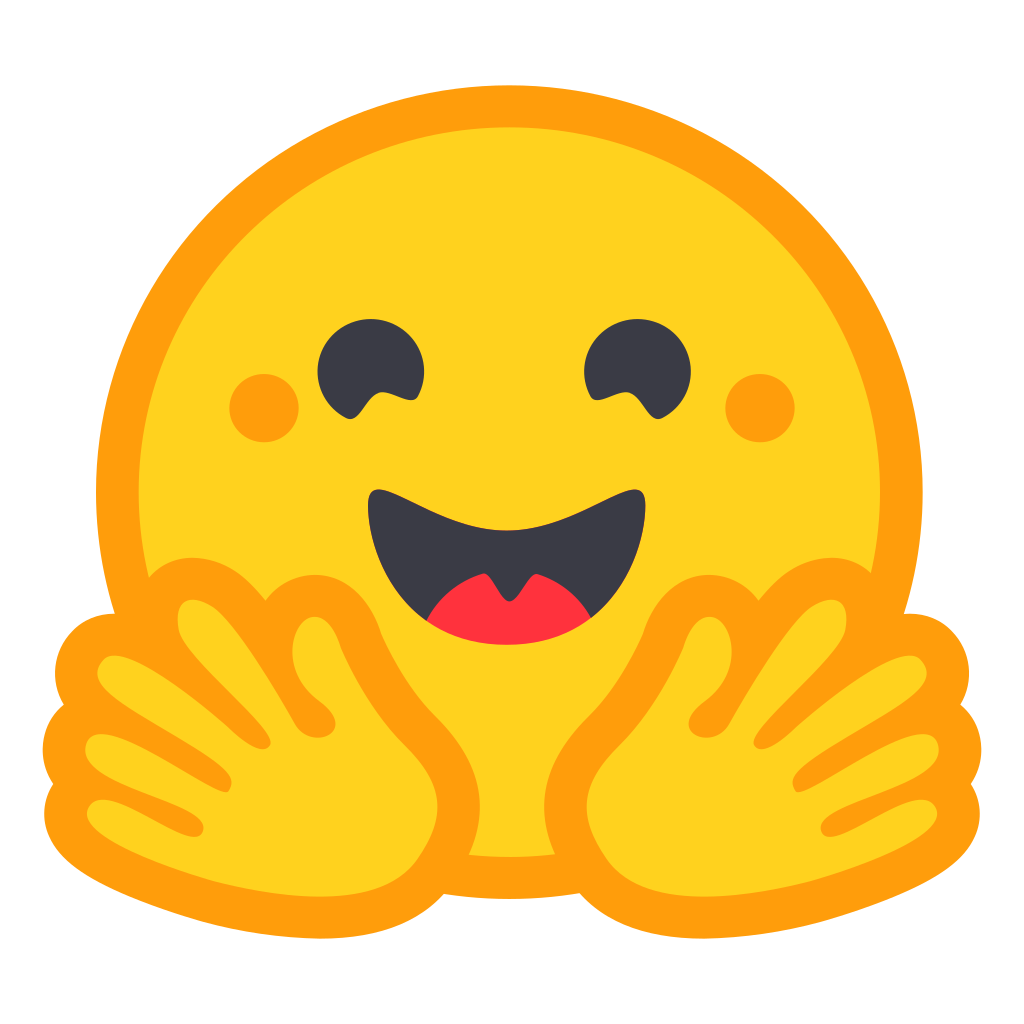}}%
}

\title{T-pro 2.0: An Efficient Russian Hybrid-Reasoning Model and Playground}

\author{Gen-T Team \\
  T-Tech, Moscow, Russia \\
   \small{
   \textbf{Correspondence:} \href{mailto:anatolii.s.potapov@gmail.com}{anatolii.s.potapov@gmail.com}
 }
  }

\begin{document}
\maketitle

\begin{abstract}
We introduce \textit{T-pro 2.0}, an open-weight Russian LLM for hybrid reasoning and efficient inference.
The model supports direct answering and reasoning-trace generation, using a Cyrillic-dense tokenizer and an adapted EAGLE speculative-decoding pipeline to reduce latency. To enable reproducible and extensible research, we release the model weights, the \textit{T-Wix} 500k instruction corpus, the \textit{T-Math} reasoning benchmark, and the EAGLE weights on Hugging Face. These resources allow users to study Russian-language reasoning and to extend or adapt both the model and the inference pipeline. A public web demo exposes reasoning and non-reasoning modes and illustrates the speedups achieved by our inference stack across domains.
T-pro 2.0 thus serves as an accessible open system for building and evaluating efficient, practical Russian LLM applications.

\hficon\ \href{https://huggingface.co/collections/t-tech/t-pro-20}
             {hf.co/collections/t-tech/t-pro-20}
\end{abstract}

\section{Introduction}
Large Language Models (LLMs) have progressed from basic text generation to systems capable of multi-step reasoning and efficient inference. Recent foundation models show that reasoning-oriented training~\citep{deepseekr1,yang2025qwen3technicalreport} and improved decoding methods~\citep{chen2023acceleratinglargelanguagemodel,Li2024EAGLE} can substantially boost both accuracy and speed.

In the Russian open-source space, progress remains limited. Most strong models are closed and accessible only through APIs~\citep{minkin2025giga,zmitrovich2023family}, while open models are typically small adaptations of multilingual systems~\citep{nikolich2024vikhr}. There is no unified ecosystem for studying Russian-language reasoning: high-quality evaluation sets are scarce, and, to the best of our knowledge, there are currently few public demos that let users compare direct answering and step-by-step reasoning, inspect inference-time optimizations, or observe how decoding speed impacts user experience.

To address these gaps, we introduce \textit{T-pro 2.0}, an open-weight Russian LLM for hybrid reasoning and an interactive demo platform. The model supports two complementary modes---direct answering and explicit reasoning traces---enabling applications to balance speed and accuracy within a single deployed system.

Our training setup combines a Cyrillic-dense tokenizer derived from Qwen3~\citep{yang2025qwen3technicalreport}, large-scale instructional midtraining, supervised fine-tuning focused on both reasoning and non-reasoning, preference optimization, and an adaptation of EAGLE-style speculative decoding~\citep{Li2024EAGLE} to accelerate Russian-language inference. 
To sum up, our main contributions are:
\begin{itemize}
    \item \textit{T-pro 2.0}, an open-weight Russian hybrid-reasoning LLM with improved inference efficiency via an optimized Cyrillic tokenizer and EAGLE-style speculative decoding.
    \item \textit{T-Wix}, the largest open Russian hybrid-reasoning SFT dataset to date (\(\approx500\text{k}\) samples) covering general instruction following, long-context tasks, and teacher-generated reasoning traces.
    \item \textit{T-Math}, a benchmark of Russian high-school olympiad-level mathematics problems for curriculum-aligned reasoning evaluation.
    \item An interactive web demo that exposes \textit{T-pro 2.0} as a research-oriented live system\footnote{The interactive demonstration interface is publicly accessible at \url{http://t-pro-2-0.streamlit.app}}\textsuperscript{,}\footnote{The \href{https://youtu.be/JcpMl5iSWC0}{web demo video} is available on YouTube.}
    , enabling side-by-side comparison of reasoning and non-reasoning modes, running tasks from our datasets and benchmarks, and viewing telemetry for inference-time optimizations.
\end{itemize}

All model-related components (T-pro~2.0, EAGLE weights, and the T-Math benchmark) are released under the Apache-2.0 license, while the T-Wix corpus is released under the ODC-By open data license.
\section{Related Work}
\begin{figure*}[!htbp]
    \centering
    \includegraphics[width=\textwidth]{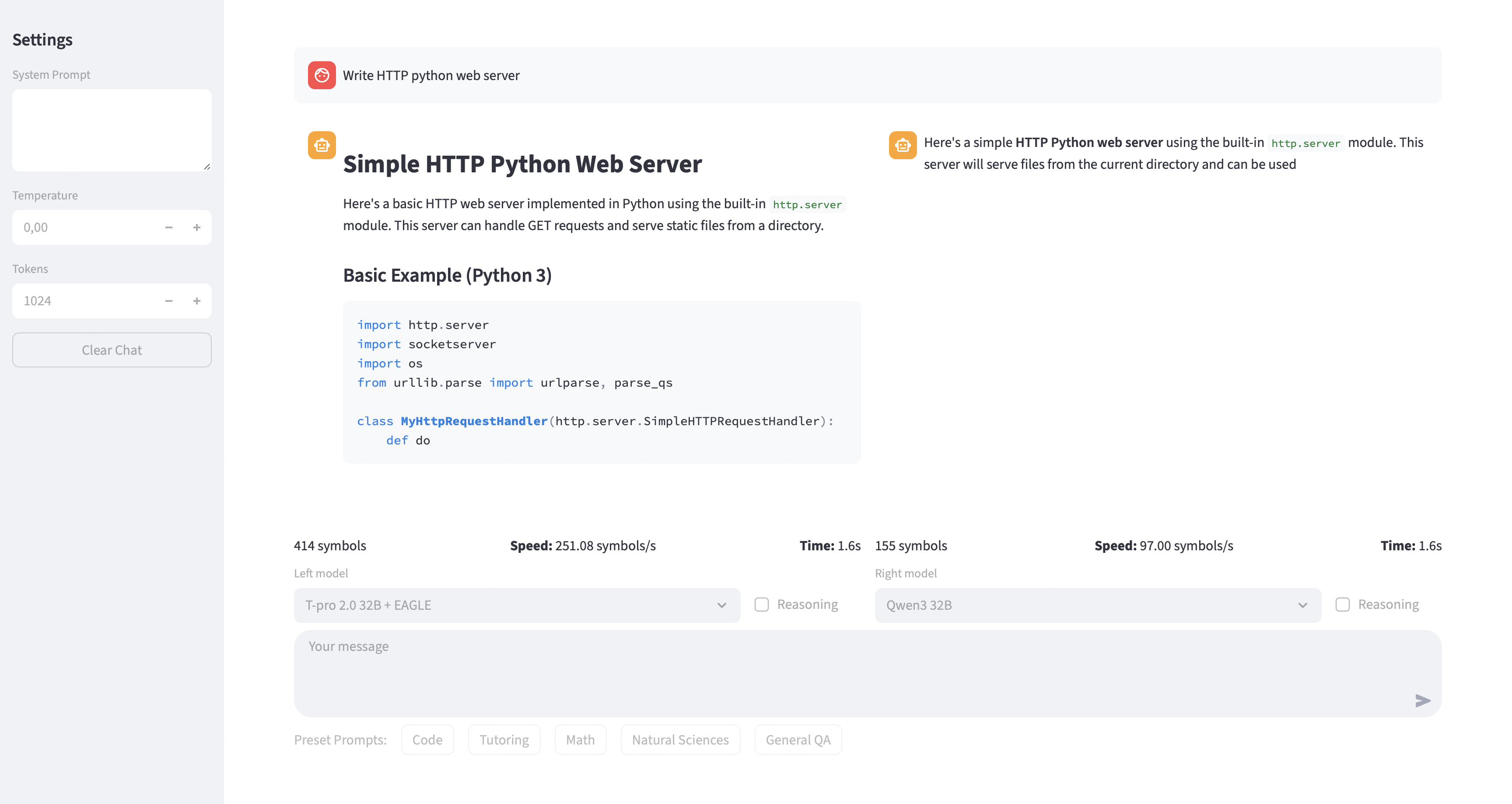}
    \caption{Screenshot of the system demo of the T-pro 2.0 EAGLE. }
    \label{fig:demo}
\end{figure*}
The development of Russian LLMs primarily follows two tracks: monolingual pre-training and adaptation of multilingual models. Early decoder-only baselines like ruGPT~\cite{kuratov2019russian, zmitrovich2023family} and commercial systems such as YandexGPT\footnote{\url{https://ya.ru/ai/gpt}} and GigaChat~\cite{minkin2025giga} focus on Russian-centric pre-training. While achieving promising results on Russian benchmarks, early versions face a capability gap compared to leading multilingual LLMs like Qwen~\cite{yang2024qwen2} and Llama~\cite{dubey2024llama3}.

To mitigate these limitations, \textit{T-pro 1.0}\footnote{\url{https://huggingface.co/t-tech/T-pro-it-1.0}} adopts a continued pre-training strategy on large-scale Russian corpora, reaching state-of-the-art results on MERA~\cite{fenogenova2024mera} among open Russian models. Its release aligns with a broader shift toward strengthening open-source Russian LLMs, alongside projects such as Saiga~\cite{gusev2023saiga}, RuAdapt~\cite{tikhomirov2024facilitating}, and Vikhr~\cite{nikolich2024vikhr}. These works emphasize the value of mitigating English-centric tokenizer limitations~\cite{petrov2024language} and extending pre-training on Russian data. This direction continues to grow: although YandexGPT-5-Lite\footnote{\url{https://huggingface.co/yandex/YandexGPT-5-Lite-8B-pretrain}} is a fully pre-trained model rather than an adaptation, its recent open release further expands the set of publicly available Russian foundation models.

\paragraph{Russian Instruction Datasets.}
Existing Russian instruction datasets vary in provenance and domain coverage. Saiga~\cite{gusev2023saiga} applies self-instruct~\cite{wang2023selfinstruct} pipelines producing ru\_turbo\_saiga, GrandMaster-PRO-MAX~\cite{nikolich2024vikhr} aggregates sources across coding and general knowledge, and RuAdapt~\cite{tikhomirov2024facilitating} combines translated and native Russian samples. However, these datasets are usually small and contain few reasoning-intensive tasks.

\paragraph{Efficient Inference.}
Speculative decoding accelerates autoregressive inference~\cite{Leviathan2023}. EAGLE~\cite{li2024eagle2} uses a lightweight head to generate draft token trees verified in parallel, achieving 2--3$\times$ speedup. Multi-Token Prediction (MTP)~\cite{gloeckle2024better} trains models to predict multiple tokens simultaneously and is deployed successfully in DeepSeek-V3~\cite{deepseekai2025deepseekv3technicalreport}. GigaChat models~\cite{minkin2025giga} also adopt MoE architecture for increased efficiency on training and inference stages. Speculative decoding remains underexplored for general-purpose Russian LLMs,
with few publicly documented deployments.
\section{T-pro 2.0}
\subsection{System and Demonstration Description}
\label{ssec:demo}

We provide a public web demo of T-pro~2.0  that exposes the model as an
interactive hybrid-reasoning assistant and makes our inference
optimizations directly observable.
The service is stateless and does not store user prompts or completions.
The interface supports multi-turn chat in Russian and English and
side-by-side comparison with baseline models (by default
Qwen3-32B-Instruct), allowing users to inspect both answers and
reasoning traces under identical serving conditions.
The demo currently supports text-only interactions and does not perform
additional server-side content filtering beyond what is built into the underlying models.

\paragraph{Architecture.}
The demo is a single-page web application backed by a lightweight
Python HTTP server.
The server exposes a simple JSON API, attaches configuration options
(model, decoding mode, generation parameters) received from the UI, and
forwards requests to two serverless SGLang endpoints~\citep{Gu2024SGLang}.
Each endpoint runs on a single NVIDIA H100 GPU: one hosts T-pro~2.0 with
an EAGLE-style speculative decoding pipeline (draft head + 32B verifier),
and the other hosts the Qwen3-32B baseline with standard autoregressive
decoding.
The deployment is tuned for interactive use and supports around 20 concurrent users per model while keeping per-request latency low.

\paragraph{User interface and functionality.}
Figure~\ref{fig:demo} shows the main layout.
The central comparison view presents parallel completions from two
systems.
For each side, users can independently choose between standard and reasoning modes.
Outputs are streamed token by token, making differences in latency,
verbosity, and reasoning structure directly visible.
A control panel above the chat area lets users select models, toggle
reasoning per model, and adjust decoding parameters such as temperature,
maximum length, and sampling options.
All decoding settings used for a given interaction are displayed in the UI, so that qualitative comparisons can be reproduced outside the demo.
A typical interaction consists of selecting a preset prompt (or entering
a custom query), choosing reasoning and generation settings, and launching both models to compare their outputs and telemetry.

To support systematic probing, the interface provides a small library of
predefined prompts grouped by domain (general questions, math and science,
code, etc.).
Several presets are derived from our evaluation suites, including
T-Math and other Russian reasoning benchmarks, so that users can quickly
examine T-pro~2.0 on challenging tasks without reconstructing
benchmark-style prompts.

\paragraph{Performance telemetry and usage patterns.}
A telemetry panel reports, for every request and model, the number of
generated tokens, end-to-end latency, streaming throughput in tokens per
second, and the acceptance ratio of speculative tokens for T-pro~2.0.
Relating these statistics to the visible outputs illustrates how
speculative decoding affects both accuracy and perceived responsiveness
for short conversational turns and long reasoning traces, complementing
the benchmark results in Section~\ref{sec:evaluation}. 

\subsection{Training recipe}
\label{ssec:training}
This section describes the T-pro 2.0 training pipeline, integrating tokenizer adaptation, instructional midtraining, general post-training, and EAGLE-based speculative decoding.
At all stages,
we perform MinHash deduplication against benchmarks to prevent data leakage.

\paragraph{Cyrillic-dense tokenizer}
We address the systematic under-tokenization of Russian in multilingual models by replacing 34k low-frequency non-Cyrillic tokens in the Qwen3~\cite{yang2025qwen3technicalreport} 
vocabulary with Cyrillic ones while keeping the total size fixed.

To build the expansion set, we extract 35.7k candidate tokens containing at least one Cyrillic character from four donor tokenizers (Qwen3, RuAdapt ~\cite{tikhomirov2024facilitating}
, cl100k\_base~\cite{openai2024gpt4ocard}, MGPT~\cite{shliazhko2023mgptfewshotlearnersmultilingual}). For each candidate, we evaluate its decomposition under the current merge graph and iteratively add those merges required to make two-piece decompositions fully reachable. Four refinement passes make approximately $95\%$ of candidates reachable. Tokens containing Cyrillic, pure Latin tokens, punctuation, and all 1–2-symbol units are preserved, while the 34k removed tokens are selected via log-smoothed frequency scoring on the midtraining mix.
\begin{table}[!htbt]
    \centering
    \resizebox{\columnwidth}{!}{%
    \begin{tabular}{cccc}
        \toprule
        
        \textbf{T-pro} & \textbf{Qwen3} &
        \textbf{GigaChat}$^\dagger$ &
        \textbf{Ruadapt-Qwen3} 
        \\
        \midrule
        \textbf{2.71} & 3.63 & 2.89 & 3.26 
        \\
        \bottomrule
    \end{tabular}%
    }
    \caption{Average tokens per word on Wikipedia for eight Cyrillic languages (ru, uk, be, bg, sr, mk, kk, ky). Lower values indicate more efficient segmentation of Cyrillic text. $^\dagger$Indicates \url{https://huggingface.co/ai-sage/GigaChat-20B-A3B-instruct} model.
    }
    \label{tab:cyrillic-tokenizers-comparison_main}
\end{table}
 Our modification yields substantial compression gains: on Russian Wikipedia, the share of Russian words tokenized into at most two tokens rises from 38\% to 60\% (see Table~\ref{tab:ru-en-tokenization}), and Tables~\ref{tab:cyrillic-tokenizers-comparison_main},~\ref{tab:cyrillic-tokenizers-comparison} further demonstrates that this improvement generalizes consistently across eight Cyrillic languages, with all of them exhibiting shorter average segmentations under our tokenizer. A full set of tokenizer evaluation metrics is provided in Appendix \ref{app:tokenizer}.

\paragraph{Instructional midtraining}
To adapt the \textit{Qwen3-32B}  model
to the new dense Russian tokenizer and enhance reasoning, we employ an intermediate stage on 40B tokens drawn from curated open-source instructions, synthetic tasks, and parallel corpora. The mixture is dominated by Russian (49\%) and English (36\%) text; in terms of domains, it is dominated by Reasoning (34.6\%), General QA (28.8\%), and Math (16.2\%), supplemented by grounded synthetic Question Answering (QA), code, and real user dialogues. The data mixture undergoes rigorous domain-specific local~sensitive hashing (LSH) deduplication and InsTag-based semantic deduplication~\cite{lu2024instag, abbas2023semdedup} to balance diversity. To ensure high quality and stylistic consistency, all assistant responses are regenerated using Qwen3-235B-A22B teacher. Training utilizes a 32k context window, stabilizing the model for downstream supervised fine-tuning (SFT).
Ablations show that the instruct-only midtraining outperforms mixtures retaining the fraction of raw pre-training data on reasoning tasks, improving ruAIME 2024~\cite{ruAIME-2024} from 0.60 to 0.67. Separately, 8B-scale experiments confirm the tokenizer transition, with the T-pro tokenizer reaching a higher MERA~\cite{fenogenova2024mera} macro-average (0.574) than the original Qwen3 tokenizer (0.560). Full details and ablations are provided in Appendix~\ref{sec:app_middle_trainig}.

\paragraph{Reward Model (RM) construction}
To support the T-pro 2.0 post-training pipeline, a dedicated reward model is trained (see Appendix~\ref{rm_training}). The RM is initialized from Qwen3-32B with a scalar regression head and trained with a Bradley--Terry preference objective on sequences up to 32K tokens using Ulysses-style sequence parallelism. Synthetic preference data are generated using knockdown tournaments over completions from multiple instruct- and reasoning-oriented model groups of different scales, substantially reducing the number of pairwise evaluations relative to an exhaustive pairwise scheme. For each instruction, completion pairs are judged, pairs with positional bias are discarded, and transitive tournament relations are added to improve preference coverage. To assess downstream performance, we design an Arena-Hard Best-of-$N$ benchmark based on the $\Delta_{\text{BoN}}$ (best@N -- worst@N) metric, on which our RM outperforms existing open-source reward models; full details are provided in Appendix~\ref{rm_analysis}.

\paragraph{General Post-Training}

The T-pro 2.0 post-training pipeline is implemented through general and reasoning SFT, and on-policy Direct Preference Optimization (DPO), with all filtering procedures detailed in Appendices~\ref{app_twix_sft}-\ref{pref_tuning}. 

For the general part of the T-Wix SFT dataset, approximately $14M$ raw instructions from open-source corpora are reduced to 468k samples using deduplication, a multi-stage filtering pipeline, and domain/complexity balancing across six domains (Math, Code, Science, General Instruction, General Knowledge, Writing) and three difficulty tiers (School, Student, Professor). Each instruction is expanded with 8 candidate completions generated by Qwen3-235B-A22B or DeepSeek-V3~\cite{deepseekai2025deepseekv3technicalreport} and then passed through an RM-guided selection step. The resulting mixture is low-noise, domain-balanced, and predominantly Russian, with approximately 10\% English data retained to preserve bilingual competence.

For the reasoning component, approximately $30$K samples are drawn from a 450k English pool covering general reasoning, mathematics, natural sciences, and code. After translation and deduplication, candidate solutions are generated by the Qwen3-235B-A22B teacher model and a midtraining student checkpoint and then filtered via RM-based rejection. For verifiable tasks, the highest-scoring factually correct teacher output is selected; for open-ended tasks, the shortest valid trace among the teacher’s top RM-ranked candidates is chosen.

DPO is performed on 100k instructions sampled from the T-Wix dataset, with a 90/10 general-to-reasoning ratio. For each instruction, 16 on-policy completions are RM-scored, and one high-contrast preference pair (best vs. worst) is formed, so that observed failure modes are directly targeted and alignment is improved without the overhead of online RL.

\paragraph{Speculative Decoding}

We integrated an EAGLE-based speculative decoding module into T-pro 2.0, where a lightweight draft model proposes candidate tokens that are verified by the frozen 32B target model. The draft model consists of a single Llama-2-based decoder layer with an FR-Spec component~\cite{zhao-etal-2025-fr}, trained on hidden-state reconstruction (smoothed $L_1$) and token distribution alignment (KL divergence) losses. During inference, we employ EAGLE-2's dynamic draft-tree mechanism via SGLang. As shown in Tables~\ref{tab:aggregated_results_no_tps} and~\ref{tab:mmlu_aggregated}, at temperature 0.8 the module achieves an average speedup of $1.85\times$ in standard mode, showing similar speedups for both standard and reasoning modes. STEM domains consistently outperform humanities ($1.99\times$ vs $1.62\times$), due to more predictable token distributions in technical content. See Appendix~\ref{app:speculative_decoding} for training pipeline details.

\begin{table}[!htbp]
\centering
\small
\resizebox{\columnwidth}{!}{%
\begin{tabular}{lcccc}
\toprule
\multirow{2}{*}{\textbf{Benchmark}} & \multicolumn{2}{c}{\textbf{Speedup}} & \multicolumn{2}{c}{\textbf{Acceptance Length}} \\
\cmidrule(lr){2-3} \cmidrule(lr){4-5}
 & \textbf{Standard} & \textbf{Reasoning} & \textbf{Standard} & \textbf{Reasoning} \\
\midrule
ruMT-Bench\textsuperscript{1} & 1.79 & 1.69 & 3.31 & 3.10 \\
ruAlpaca\textsuperscript{2} & 1.61 & 1.57 & 2.94 & 2.85 \\
ruCodeEval\textsuperscript{3} & 2.15 & 1.84 & 3.93 & 3.34 \\
T-Math\textsuperscript{4} & -- & 2.25 & -- & 4.01 \\
\midrule
\textbf{Average} & \textbf{1.85} & \textbf{1.83} & \textbf{3.39} & \textbf{3.33} \\
\bottomrule
\end{tabular}%
}
\caption{Aggregated T-pro-2.0-eagle performance at temperature 0.8. Comparison of relative speedup and average acceptance length for the standard and reasoning modes.
\textsuperscript{1}\citet{ru-mt-bench}, \textsuperscript{2}\citet{ru-alpaca-eval}, \textsuperscript{3}\citet{fenogenova2024mera}, \textsuperscript{4}\citet{T-math}
}
\label{tab:aggregated_results_no_tps}
\end{table}
\begin{table}[!htbp]
\small
\centering
\begin{tabular}{lcc}
\toprule
\textbf{Domain Category} & \textbf{Speedup} & \textbf{Acceptance Length} \\
\midrule
STEM \& Business$^{\dagger}$ & 1.99 & 3.57 \\
Social \& Humanities$^{\ddagger}$ & 1.62 & 2.88 \\
\midrule
\textbf{Average} & \textbf{1.79} & \textbf{3.19} \\
\bottomrule
\end{tabular}%
\caption{Aggregated acceleration results on ruMMLU-Pro. 
{\small $^{\dagger}$Includes Math, Chem, Eng, Bus, Phys, CS. 
$^{\ddagger}$Includes Econ, Bio, Health, Psych, Phil, Hist, Law.}}
\label{tab:mmlu_aggregated}
\end{table}
\begin{table*}[!htbt]
\centering
\small
\setlength{\tabcolsep}{2pt}
\resizebox{\textwidth}{!}{%
\begin{tabular}{@{}lccccccc@{}}
\toprule
\multirow{2}{*}{\textbf{Model}} & \multirow{2}{*}{\textbf{MERA}} & \multirow{2}{*}{\textbf{MaMuRAMu}} & \multirow{2}{*}{\textbf{ruMMLU-Pro}} & \textbf{Arena} & \textbf{WildChat} & \multicolumn{2}{c}{\textbf{Arena Hard 2}} \\
 & & & & \textbf{Hard Ru} & \textbf{Hard Ru} & \textbf{HP} & \textbf{CW} \\\midrule
\textit{Open Source Models (27B-32B class)} &  &  &  &  &  &  &  \\
\cellcolor{blue!10}\textbf{T-pro 2.0 (Ours)} & \cellcolor{blue!10}\textbf{0.66} & \cellcolor{blue!10}\textbf{0.851} & \cellcolor{blue!10}\textbf{0.697} & \cellcolor{blue!10}\textbf{91.1 / 90.36} & \cellcolor{blue!10}\textbf{72.6 / 76.4} & \cellcolor{blue!10}\textbf{53.5 / 46.2} & \cellcolor{blue!10}\textbf{64.2 / 62.8} \\
Qwen3-32B & \underline{0.582} & \underline{0.833} & \underline{0.677} & \underline{83.95 / 84.66} & \underline{59.6 / 51.9} & \underline{46.4 / 32.9} & \underline{53.7 / 41.5} \\
RuadaptQwen3-32B-Instruct\textsuperscript{1} & 0.574 & 0.823 & 0.652 & 68.4 / 64.76 & 41.5 / 39.4 & 13 / 14.2 & 19.4 / 12.7  \\
Gemma 3 27B\textsuperscript{2}
& 0.577 & 0.808 & 0.665 & 82.66 & 58.4 & 23.5 & 74.7 \\
DeepSeek-R1-Distill-Qwen-32B\textsuperscript{3} & 0.508 & 0.787 & 0.537 & 34.07 / 22.83 & 12.1 / 8.7 & 7.2 / 7.2 & 5.9 / 3.5 \\
\midrule
\multicolumn{8}{l}{\textit{Open Source Larger Scale \& Proprietary Models}} \\
DeepSeek-V3 & \textbf{0.677} & \textbf{0.875} & \textbf{0.736} & 92.67 & 66.8 & 45.8 & \underline{59.9} \\
DeepSeek-R1\textsuperscript{3} & -- & -- &  -- 
& \textbf{95.74} & \textbf{90.3} & \textbf{73.6} & \textbf{90.3} \\
YandexGPT5-Pro\textsuperscript{4} & -- & -- & 0.604 & 19.13 & 12.1 & 3.8 & 2.6 \\
GigaChat 2 Max\textsuperscript{5}
& 0.67 & 0.864 & 0.649 & 61.44 & 10.1 & 8.5 & 27.1 \\
o4-mini\textsuperscript{6} (medium) & -- & -- & -- & \underline{95.63} & \underline{74.4} & \underline{67} & 49.8 \\
GPT-4o\textsuperscript{7} & \underline{0.642} & \underline{0.874} & \underline{0.714} & 85.14 & 41.4 & 20.0 & 44.2 \\  
\bottomrule
\multicolumn{8}{l}{
\textsuperscript{1}\citet{tikhomirov2024facilitating}, \textsuperscript{2}\citet{gemma3}, \textsuperscript{3}\citet{deepseekr1}, \textsuperscript{4}\url{https://ya.ru/ai/gpt},} \\
\multicolumn{8}{l}{
\textsuperscript{5}\citet{minkin2025giga}, \textsuperscript{6}\citet{openai2025o3o4mini}, \textsuperscript{7}\citet{openai2024gpt4ocard}.} \\
\end{tabular}
}
\caption{Comparison of models on Russian language understanding and dialogue benchmarks. In Arena Hard 2, subsets are Hard Prompt (HP) and Creative Writing (CW). For entries reported as $a/b$, the first value corresponds to the \textit{reasoning} setting and the second to the \textit{non-reasoning} setting. o4-mini and DeepSeek-R1 are omitted from MERA as it does not support reasoning model mode, while YandexGPT5-Pro is omitted from MERA due to licensing restrictions.
}
\label{tab:main_results2}
\end{table*}
\begin{table*}[!htbt]
\centering
\small
\setlength{\tabcolsep}{4pt}
\resizebox{\textwidth}{!}{%
\begin{tabular}{@{}lcccccccc@{}}
\toprule
\textbf{Model} & \multirow{2}{*}{\textbf{T-Math}} & \textbf{ruAIME} & \textbf{ruAIME} & \textbf{ruMATH-500} & \textbf{ruGPQA} & \multirow{2}{*}{\textbf{ruLCB}} & \textbf{Vikhr} & \textbf{Vikhr} \\
& & \textbf{2024} & \textbf{2025} &  & \textbf{Diamond} & & \textbf{Math} & \textbf{Physics} \\
\midrule
\textit{Open Source Models (27B-32B class)} & &  &  &  &  &  &  &  \\
\cellcolor{blue!10}\textbf{T-pro 2.0 (Ours)} & \cellcolor{blue!10}\textbf{0.541} & \cellcolor{blue!10}\underline{0.704} & \cellcolor{blue!10}\textbf{0.646} & \cellcolor{blue!10}\textbf{0.94} & \cellcolor{blue!10}0.591 & \cellcolor{blue!10}\textbf{0.563} & \cellcolor{blue!10}\underline{0.799} & \cellcolor{blue!10}\underline{0.51}\\
Qwen3-32B & \underline{0.529} & \textbf{0.706} & \underline{0.625} & \underline{0.938} & \underline{0.606} & \underline{0.537} & \textbf{0.809} & \textbf{0.531} \\
RuadaptQwen3-32B-Instruct & 0.444 & 0.575 & 0.450 & 0.450 & 0.591 & 0.500 & 0.528 & 0.337 \\
Gemma 3 27B & 0.208 & 0.248 & 0.231 & 0.860 & 0.439 & 0.261 & 0.548 & 0.276 \\
DeepSeek-R1-Distill-Qwen-32B & 0.254 & 0.510 & 0.402 & 0.898 & \textbf{0.631} & 0.493 & 0.462 & 0.286 \\
\midrule
\multicolumn{8}{l}{\textit{Open Source Larger Scale \& Proprietary Models}} \\
DeepSeek-V3 & 0.278 & 0.319 & 0.285 & 0.882 & 0.657 & 0.444 & 0.613 & 0.367 \\
DeepSeek-R1 & \underline{0.619} & \textbf{0.800} & \textbf{0.800} & \textbf{0.972} & \underline{0.763} & \underline{0.69} & \textbf{0.864} & \textbf{0.469} \\
YandexGPT5-Pro & 0.13 & 0.062 & 0.046 & 0.682 & 0.354 & 0.265 & 0.372 & 0.252 \\
GigaChat 2 Max & 0.142 & 0.102 & 0.062 & 0.702 & 0.475 & 0.272 & 0.372 & 0.245 \\
o4-mini (medium) & \textbf{0.634} & \underline{0.781} & \underline{0.771} & \underline{0.958} & \textbf{0.773} & \textbf{0.705} & \underline{0.834} & \underline{0.408} \\
GPT-4o & 0.106 & 0.090 & 0.069 & 0.766 & 0.510 & 0.131 & 0.372 & 0.296 \\
\bottomrule
\end{tabular}
}
\caption{Comparison of models on Russian advanced reasoning benchmarks.}
\label{tab:ru_math_results_final}
\end{table*}

\section{Evaluation}\label{sec:evaluation}
\subsection{Benchmarks}
We evaluate along three axes: factual knowledge, dialogue, and reasoning capabilities.

\paragraph{Russian common-knowledge benchmarks}
MERA, MaMuRAMu~\cite{fenogenova2024mera}, and ruMMLU-Pro~\cite{ruMMLUpro}, targeting world knowledge and logical competence.

\paragraph{Dialogue benchmarks}
Arena Hard Ru~\cite{ru_arena_hard}, Arena Hard 2 \cite{li2024crowdsourced, arenahard2024}, and WildChat Hard Ru~\cite{wildchat-hard-ru} (curated native Russian queries). WildChat uses o3-mini responses as baseline; DeepSeek-V3.1-Terminus~\cite{deepseekai2025deepseekv3technicalreport} serves as judge across all arenas and DeepSeek-V3.1~\cite{deepseekai2025deepseekv3technicalreport} for WildChat.

\paragraph{Reasoning benchmarks}
AIME 24/25~\cite{aime24}~\cite{aime25}, MATH-500~\cite{hendrycks2021math}, GPQA Diamond~\cite{rein2023gpqa}, Vikhr Math/Physics ~\cite{Kuleshov2025DOoM}, and LiveCodeBench v4\_v5~\cite{jain2024livecodebench}. English benchmarks are professionally localized (ruAIME, ruMATH-500, ruGPQA, ruLCB \cite{ru_reasoning_benchmarks}). Vikhr benchmarks use Math-Verify scoring~\footnote{\url{https://github.com/huggingface/Math-Verify}}.

\paragraph{T-Math benchmark}
We introduce T-Math—331 problems from All-Russian and Moscow olympiads (1998–2025), automatically extracted and human-verified. Details are provided in Appendix~\ref{app:tmath}.

\subsection{Results}

\paragraph{General knowledge and dialogue abilities}
Table~\ref{tab:main_results2} shows the results on Russian general-knowledge and dialogue benchmarks. T-pro 2.0 performs consistently well across all evaluations, scoring 0.66 on MERA and 0.697 on ruMMLU-Pro. These numbers put it close to GPT-4o (0.714) and above Russian-adapted baselines such as RuadaptQwen3-32B-Instruct (0.652).

On dialogue tasks, the model reaches 91.1 on Arena Hard Ru and 72.6 on WildChat Hard Ru, outperforming all open-source systems and most proprietary ones. On Arena Hard 2, T-pro 2.0 scores 53.5 on Hard Prompts and 64.8 on Creative Writing, showing that it reliably follows instructions across different task types. These results directly reflect the structure of the T-Wix corpus, which mixes general instruction-following with long-context tasks and distilled reasoning traces from stronger teacher models.

\paragraph{Reasoning capabilities}
Table~\ref{tab:ru_math_results_final} summarizes performance on T-Math and localized reasoning benchmarks. On T-Math, the model scores 0.541, indicating strong performance on original olympiad-style Russian problems. On ruAIME 2024 and 2025 it reaches 0.704 and 0.646, sharply outperforming DeepSeek-V3 (0.319/0.285), GPT-4o (0.090/0.069) and all proprietary Russian models. Results on ruMATH-500 (0.94) and Vikhr Math (0.799) further confirm the model’s ability to perform mathematical reasoning in Russian under varied setups. These results also show that T-Math is a challenging benchmark that reveals meaningful performance differences that are obscured by translated or adaptation-based alternatives.

Crucially, the Russian-focused training does not hurt English performance. As shown in Table~\ref{tab:en_reasoning2}, \mbox{T-pro~2.0} remains competitive on English reasoning benchmarks, with 0.765 on AIME 2024, 0.966 on MATH-500, and 0.556 on LiveCodeBench, on par with or better than other open-source models of similar scale. A detailed breakdown of English results is provided in Appendix~\ref{en_results_benches}.
\section{Conclusion}

We present \textit{T-pro 2.0}, an open-weight Russian language model tailored for hybrid reasoning and efficient inference. The combination of a Cyrillic-dense tokenizer, a reasoning-focused midtraining stage, and an adapted EAGLE-style speculative decoding pipeline allows the model to deliver strong performance on Russian tasks without increasing model size and without notable degradation on English benchmarks. Along with the model, we release \textit{T-Wix}, a large-scale SFT dataset enriched with reasoning traces, and \textit{T-Math}, a benchmark designed to probe mathematical and analytical abilities in Russian.

These results point to two broader takeaways. First, careful, targeted adaptation of strong multilingual backbones remains a practical and reproducible route for building high-quality models for languages with limited resources. Second, tokenizer design and inference optimization are not optional details but key components for deploying reasoning-capable models beyond English. We expect the released models, datasets, evaluation code, and public demo to support research on Russian-language reasoning LLMs and to contribute to more transparent and consistent evaluation practices in this area.
\section*{Ethical Statement}

\paragraph{Possible Misuse.}
Our work may enable generation of misleading, offensive, or otherwise harmful content if deployed without appropriate safeguards. We do not support applications that restrict access to information, facilitate disinformation, target individuals or groups, or automate harmful actions. To mitigate these risks, we apply toxicity and safety filtering during post-training and provide usage guidelines for responsible deployment.

\paragraph{Biases and Data Quality.}
The datasets used for pre-training and fine-tuning contain publicly available Russian and English text, which may include stereotypes, factual inaccuracies, or cultural biases. While automated filtering and manual checks are applied, residual biases may remain. We recommend additional evaluation when transferring the model to domains or communities that are underrepresented in the training data.

\paragraph{Human Subjects and Privacy.}
This work does not involve human-subjects research or the collection of personally identifiable information. All data sources comply with their respective licenses and usage policies.

\section*{Limitations}

Despite the strong performance of T-pro 2.0, several limitations should be acknowledged, which are planned to be addressed in future work.

\textbf{Limited Agentic Capabilities} No dedicated improvements for tool use or agentic behavior were incorporated into the model. Optimizations for function calling or complex multi-turn interactions were not performed, and as a result, performance in these areas is expected to be comparable to or slightly below that of the base Qwen3-32B model. Enhancements in this direction are prioritized for future development.

\textbf{Offline-Only Reinforcement Learning}
Model alignment was conducted exclusively through offline DPO.
Online reinforcement learning methods such as PPO~\citep{schulman2017proximalpolicyoptimizationalgorithms}
or GRPO~\citep{shao2024deepseekmathpushinglimitsmathematical} were not employed.
Although DPO is computationally efficient, the absence of interactive feedback
may limit robustness and lead to performance degradation on out-of-domain tasks.

\textbf{Unverified Long-Context Performance} All training stages for T-pro 2.0 were carried out with a fixed context window of 32k tokens, consistent with the base Qwen3-32B configuration. While support for up to 128k tokens is theoretically enabled via RoPE scaling, the model's ability to maintain coherence and retrieve information over such extended contexts has not been empirically validated.

\textbf{Reproducibility Issues} Full reproducibility is restricted by the use of proprietary datasets in midtraining and the DPO stage. However, the curated SFT dataset is being released to support and encourage further research, particularly in the development of high-quality Russian-language language models.

\section*{Author Contributions}
\begin{itemize}
    \item \emph{Administration and Supervision}: Anatolii Potapov
    \item \emph{Training Pipelines Team}: German Abramov, Pavel Gein
    \item \emph{Post-training Team}: Dmitrii Stoianov, Olga Tsymboi, Danil Taranets, 
    Ramil Latypov, Almaz Dautov, Dmitry Abulkhanov
    \item \emph{Inference Team}: Vladislav Kruglikov, Nikita Surkov
    \item \emph{Evaluation Team}: Mikhail Gashkov, 
    Viktor Zelenkovskiy, Artem Batalov, Alexandr Medvedev
\end{itemize}


\bibliography{custom}

\newpage
\appendix
\newpage
\
\newpage
\begin{table*}[!htbp]
\small
\centering
\begin{tabular}{l l l l}
\toprule
Resource & Type & Location & License \\
\midrule
T-pro 2.0 & Model & \url{https://huggingface.co/t-tech/T-pro-it-2.0} & Apache-2.0 \\
EAGLE weights & Model component & \url{https://huggingface.co/t-tech/T-pro-it-2.0-eagle} & Apache-2.0 \\
T-Math & Benchmark dataset & \url{https://huggingface.co/datasets/t-tech/T-math} & Apache-2.0 \\
T-Wix 500k & Instruction corpus & \url{https://huggingface.co/datasets/t-tech/T-Wix} & ODC-By \\
\bottomrule
\end{tabular}
\caption{Released resources and licenses.}
\label{tab:resources}
\end{table*}

\section{Released Resources and Licenses}
\label{app:resources}

All released components use permissive, research-friendly licenses. The T-pro 2.0 model, its EAGLE draft weights, and the T-Math benchmark are distributed under Apache-2.0, allowing broad academic and commercial use. The T-Wix 500k corpus is released under ODC-By. Full license details appear in Table \ref{tab:resources}.

\section{Tokenizer adaptation statistics}
\label{app:tokenizer}

\begin{figure*}[!htbp]
    \centering
    \begin{subfigure}{0.47\textwidth}
        \includegraphics[width=\textwidth]{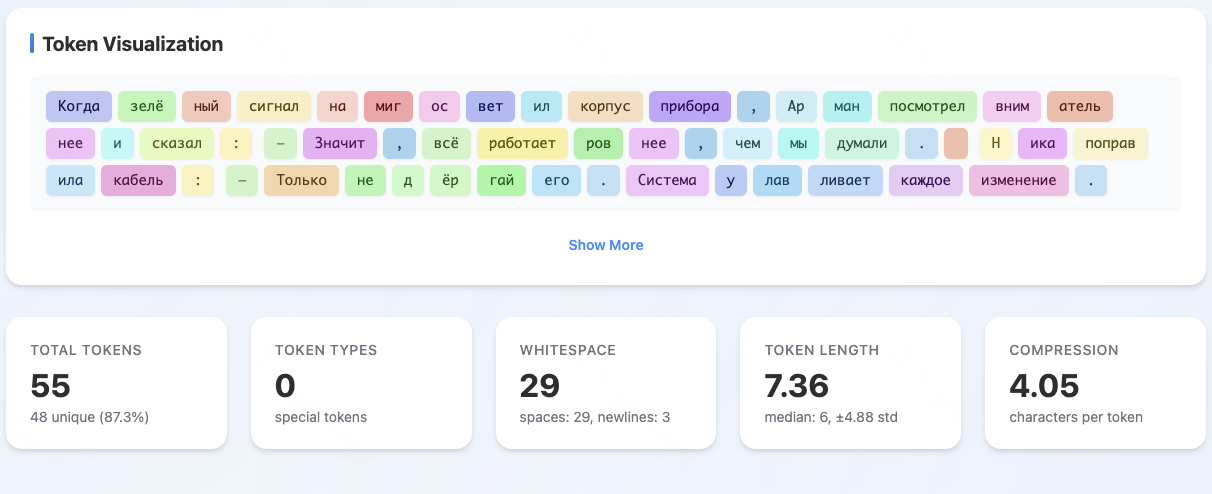}
        \caption{T-pro}
    \end{subfigure}~
    \begin{subfigure}{0.47\textwidth}
        \includegraphics[width=\textwidth]{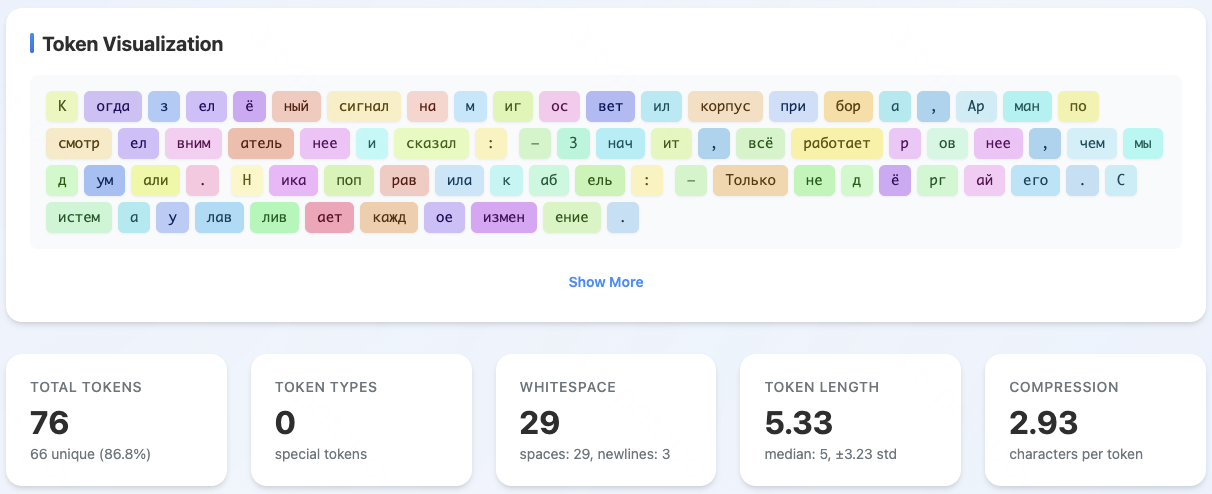}
        \caption{Qwen3}
    \end{subfigure}
    \\
    \begin{subfigure}{0.47\textwidth}
        \includegraphics[width=\textwidth]{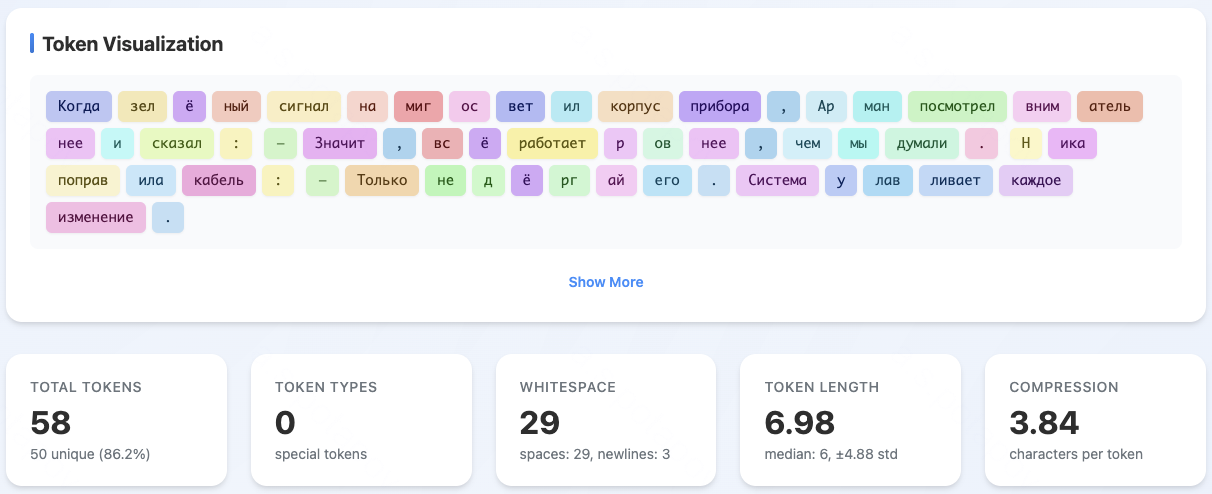}
        \caption{Ruadapt-Qwen3}
    \end{subfigure}
    ~
    \begin{subfigure}{0.47\textwidth}
        \includegraphics[width=\textwidth]{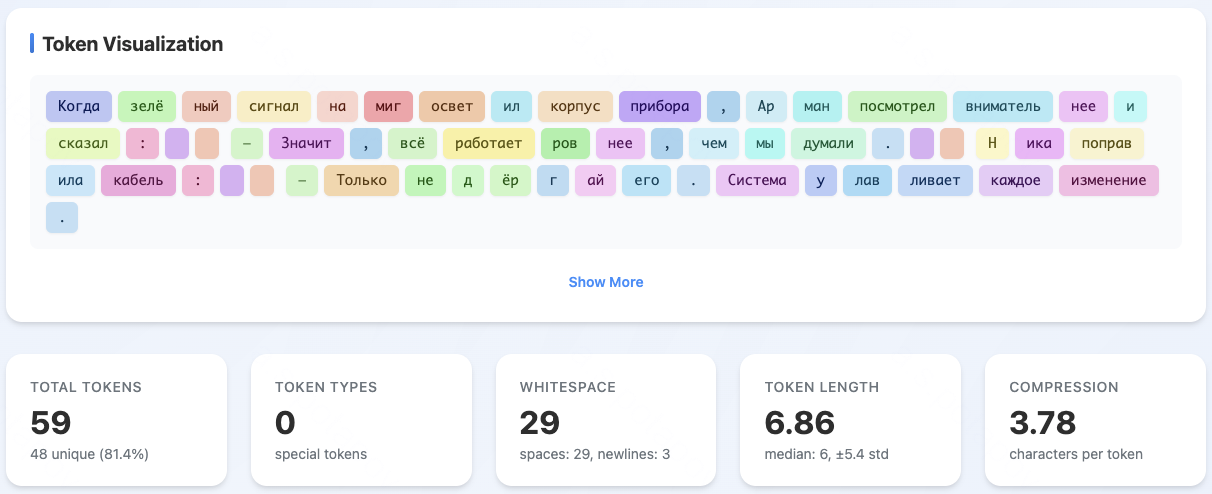}
        \caption{GigaChat}
    \end{subfigure}
    \caption{Qualitative comparison of Russian tokenization. A 220-character text is tokenized by T-pro 2.0, the original Qwen3 tokenizer, and other Cyrillic-optimized models. T-pro 2.0 encodes the text into just 55 tokens compared to 76 for Qwen3, demonstrating superior compression efficiency.}
    \label{fig:tokenization-example}
\end{figure*}

\paragraph{Russian and English corpora.}
Table~\ref{tab:ru-en-tokenization} reports tokenization statistics for Russian
and English on both Wikipedia and our in-domain SFT corpus (T-Wix).
For Russian, the Cyrillic-dense T-pro 2.0 tokenizer substantially reduces
the average number of tokens per word and increases the share of words
segmented into at most two tokens, while English compression is essentially
unchanged.

\begin{table}[!htbp]
    \centering
    \small
    \setlength{\tabcolsep}{3pt}
    \begin{tabular}{@{}llcccc@{}}
        \toprule
        \textbf{Corpus} & \textbf{Tokenizer} &
        \textbf{\begin{tabular}[c]{@{}c@{}}tok/\\word\end{tabular}} & 
        \textbf{\begin{tabular}[c]{@{}c@{}}1 tok\\(\%)\end{tabular}} &
        \textbf{\begin{tabular}[c]{@{}c@{}}$\leq$2 tok\\(\%)\end{tabular}} & 
        \textbf{\begin{tabular}[c]{@{}c@{}}>2 tok\\(\%)\end{tabular}} \\
        \midrule
        \multicolumn{6}{@{}l}{\textit{Russian}} \\
        \quad ruWiki  & Qwen3     & 3.12 & 20.3 & 38.2 & 61.8 \\
        \quad ruWiki  & T-pro 2.0 & 2.38 & 28.7 & 60.1 & 39.9 \\
        \quad T-Wix   & Qwen3     & 2.70 & 31.8 & 52.4 & 47.6 \\
        \quad T-Wix   & T-pro 2.0 & 2.26 & 39.3 & 65.5 & 34.5 \\
        \midrule
        \multicolumn{6}{@{}l}{\textit{English}} \\
        \quad enWiki  & Qwen3     & 1.68 & 61.2 & 83.7 & 16.3 \\
        \quad enWiki  & T-pro 2.0 & 1.68 & 61.1 & 83.7 & 16.3 \\
        \bottomrule
    \end{tabular}
    \caption{Tokenization density statistics for Russian and English on Wikipedia and our SFT corpus (T-Wix). We compare the original Qwen3 and Cyrillic-dense T-pro 2.0 tokenizers. Columns show: average tokens per word (tok/
word), percentage of words tokenized into exactly 1 token, at most 2 tokens, and more than 2 tokens.}
    \label{tab:ru-en-tokenization}
\end{table}

Table~\ref{tab:cyrillic-wiki-tokenization} extends this analysis to eight
Cyrillic languages using Wikipedia. In all cases the new tokenizer reduces
tokens per word, with particularly large gains for Kyrgyz, which is poorly
served by generic multilingual tokenizers.

\begin{table}[!htbp]
  \centering
  \small
  \begin{tabular}{lcccc}
    \toprule
    & \multicolumn{2}{c}{\textbf{Tokens/Word}} & \multicolumn{2}{c}{\textbf{\% Words ($\leq$2 tok)}} \\
    \cmidrule(lr){2-3} \cmidrule(lr){4-5}
    \textbf{Lang} & \textbf{Qwen3} & \textbf{T-pro} & \textbf{Qwen3} & \textbf{T-pro} \\
    \midrule
    ru & 3.12 & 2.38 & 38.20 & 60.13 \\
    uk & 3.70 & 2.80 & 31.17 & 45.79 \\
    be & 3.97 & 2.94 & 30.15 & 41.36 \\
    bg & 2.99 & 2.35 & 43.42 & 59.60 \\
    sr & 3.26 & 2.62 & 37.65 & 51.79 \\
    mk & 3.04 & 2.41 & 42.42 & 57.19 \\
    kk & 4.60 & 3.07 & 15.30 & 37.69 \\
    ky & 4.35 & 3.09 & 21.27 & 39.93 \\
    \bottomrule
  \end{tabular}%
  \caption{Tokenization density on Wikipedia for Cyrillic languages. \textit{Tokens/Word}: average tokens per word; \textit{\% Words ($\leq$2 tok)}: percentage of words tokenized into at most 2 tokens.}
  \label{tab:cyrillic-wiki-tokenization}
\end{table}

\paragraph{Comparison with other Cyrillic-rich tokenizers.}
Finally, Table~\ref{tab:cyrillic-tokenizers-comparison} compares T-pro 2.0
against several strong Cyrillic-focused baselines.
T-pro 2.0 achieves the lowest tokens-per-word for seven out of eight languages
(ru, uk, be, bg, sr, mk, ky) and remains competitive on Kazakh,
demonstrating that our tokenizer design is competitive with specialized
alternatives.

\begin{table}[!htbp]
    \centering
    \resizebox{\columnwidth}{!}{%
    \begin{tabular}{lcccc}
        \toprule
        \textbf{Lang} &
        \textbf{T-pro} &
        \textbf{GigaChat}$^\dagger$ &
        \textbf{Ruadapt-Qwen3} &
        \textbf{gpt-oss} \\
        \midrule
        ru & \textbf{2.38} & 2.49 & 2.43 & 2.70 \\
        uk & \textbf{2.80} & 3.09 & 3.29 & 2.92 \\
        be & \textbf{2.94} & 3.32 & 3.54 & 3.03 \\
        bg & \textbf{2.35} & 2.58 & 2.50 & 2.56 \\
        sr & \textbf{2.62} & 2.97 & 3.07 & 2.73 \\
        mk & \textbf{2.41} & 2.67 & 2.70 & 2.59 \\
        kk & 3.07          & \textbf{2.67} & 4.60 & 3.11 \\
        ky & \textbf{3.09} & 3.33 & 3.97 & 3.17 \\
        \midrule
        Avg & \textbf{2.71} & 2.89 & 3.26 & 2.85 \\
        \bottomrule
    \end{tabular}%
    }
    \caption{Tokenization density (tokens/word) on Wikipedia for T-pro and other Cyrillic-rich tokenizers. Lower is better. $^\dagger$Indicates \url{https://huggingface.co/ai-sage/GigaChat-20B-A3B-instruct} model.}
    \label{tab:cyrillic-tokenizers-comparison} 
\end{table}
\section{Instructional midtraining}
\label{sec:app_middle_trainig}

\begin{figure*}[t]
    \centering
    \includegraphics[width=\textwidth]{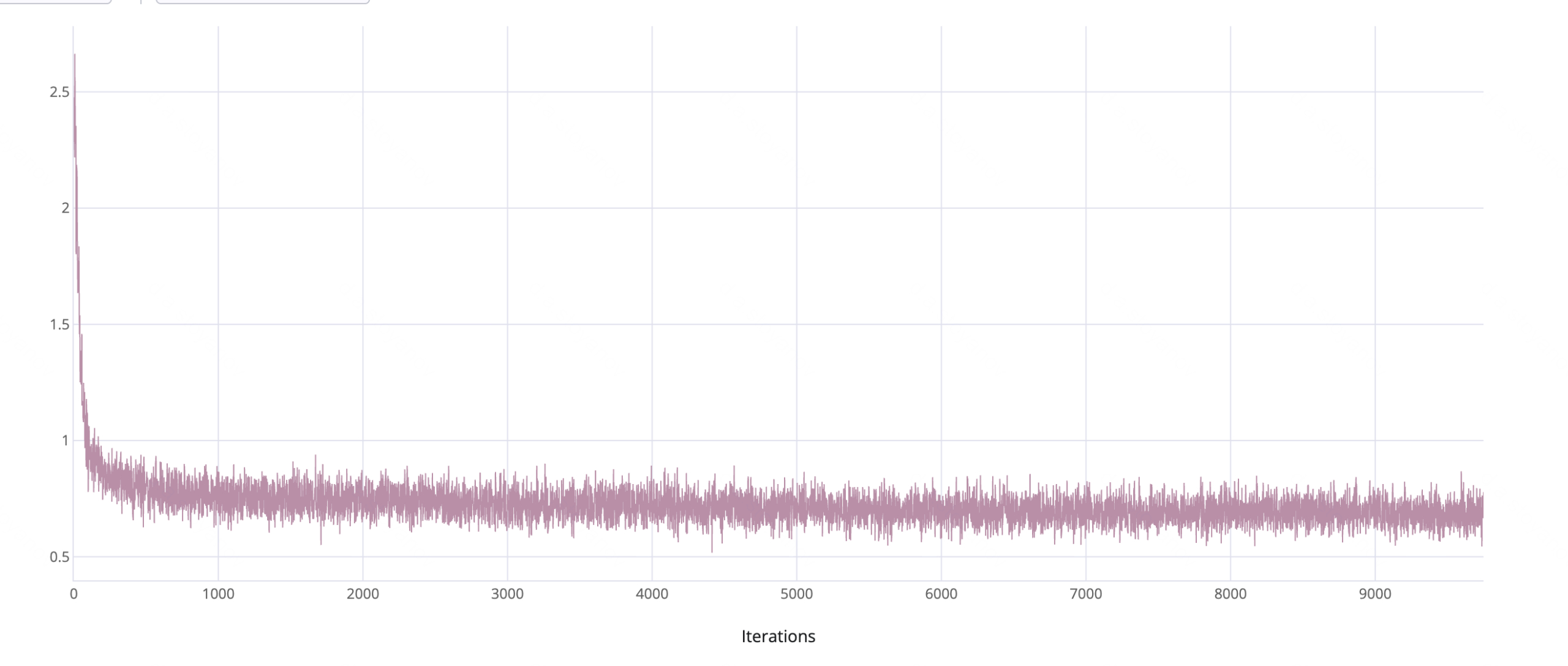}
    \caption{Midtraining training loss as a function of optimization
    steps. Loss drops steeply during the first $\sim$1k steps
    ($\sim$4B tokens) and then gradually plateaus, indicating that most
    adaptation to the new tokenizer happens early in the run.}
    \label{fig:middle-training-loss}
\end{figure*}

We employ an intermediate \emph{instructional midtraining} stage between generic
large-scale pre-training and downstream alignment.
Starting from the publicly available Qwen3-32B dense model~\cite{yang2025qwen3technicalreport},
already pre-trained on $\sim 36$T tokens, we perform continual training on
40B tokens of instruction-style data.
The goals of this stage are:
(i) adapt the model to a denser, Russian-centric tokenizer,
(ii) learn useful representations for new subword units, and
(iii) further strengthen Russian language and reasoning skills without degrading
the base model’s capabilities.

\paragraph{Training setup.}
\begin{table}[t]
    \centering
    \resizebox{\columnwidth}{!}{
    \begin{tabular}{ll}
        \toprule
        \textbf{Hyperparameter} & \textbf{Value} \\
        \midrule
        Global batch size (tokens) & 4M (128 seq $\times$ 32K) \\
        Total tokens & 40B \\
        Steps & $\approx 9{,}750$ \\
        Max context length & 32K \\
        \midrule
        Optimizer & AdamW \\
        Adam betas & $(0.9, 0.95)$ \\
        Adam $\epsilon$ & $10^{-8}$ \\
        Weight decay & $10^{-6}$ \\
        LR schedule & cosine \\
        Peak / min LR & $1\times10^{-5}$ / $1\times10^{-6}$ \\
        Warmup & 100 batches \\
        Gradient clipping & max norm 1.0 \\
        \midrule
        Precision & BF16 \\
        Parallelism & FSDP full-shard, act. checkpointing \\
        \bottomrule
    \end{tabular}
    }
    \caption{Midtraining optimization setup.}
    \label{tab:middle_training_hparams}
\end{table}

Midtraining uses a 4M-token global batch and 40B total tokens ($\approx 9{,}750$ steps). We train with AdamW using a peak learning rate of $1\times10^{-5}$ and cosine decay to $1\times10^{-6}$, with 100 warmup batches. Data are formatted in the same chat-style schema and packed up to a 32K context window without a length curriculum. Training is performed in bf16 with FSDP full-shard and activation checkpointing. Remaining hyperparameters are listed in Table~\ref{tab:middle_training_hparams}.

\paragraph{Midtraining datamix}

All midtraining samples are in instruction format.
The datamix combines
(i) public instruction datasets from the Hugging Face Hub,
(ii) web and forum data (e.g., question-answer style threads),
(iii) real user--assistant dialogues, and
(iv) synthetic instructional data and reasoning traces grounded in pre-training corpora
(books, Common Crawl, code) via a WebInstruct-style pipeline~\cite{NEURIPS2024_a4ca07aa}.
Compared to the SFT stage, the midtraining datamix is intentionally
\emph{larger and less curated}: we trade some noise for broader coverage of
tasks and domains.
All instructions are derived from public sources; internal data are used
only as anonymized targets in dialogue-style responses.

Table~\ref{tab:middle-training-datamix} reports the category-level breakdown
of the 40B-token corpus.
In terms of language, the corpus is predominantly Russian and English:
roughly 49\% of tokens are Russian (19.6B), 36\% English (14.4B),
9.3\% code (3.7B) and 5.5\% come from parallel Russian--English data (2.2B).

\begin{table}[!htbp]
    \centering
    \small
    \begin{tabular}{lcc}
        \toprule
        \textbf{Category} & \textbf{Share} & \textbf{Tokens (B)} \\
        \midrule
        Reasoning           & 34.5\% & 13.8 \\
        General             & 29.3\% & 11.7 \\
        Math                & 16.3\% & 6.5 \\
        Real chat           & 5.5\%  & 2.2 \\
        IF                  & 5.0\%  & 2.0 \\
        Grounded QA synth   & 3.8\%  & 1.5 \\
        Code                & 2.8\%  & 1.1 \\
        Forum               & 1.7\%  & 0.7 \\
        Summarization       & 0.7\%  & 0.3 \\
        ICL                 & 0.4\%  & 0.2 \\
        \bottomrule
    \end{tabular}
    \caption{Midtraining datamix by category (40B tokens total).}
    \label{tab:middle-training-datamix}
\end{table}

\paragraph{InsTag deduplication.}
To control redundancy while preserving diversity across sources, we apply
\#\textsc{InsTag}-based deduplication~\cite{lu2024instag}
\emph{independently within each component} of the datamix (reasoning,
general QA, code, etc.).
The tagger is applied only to user utterances; all tags from a multi-turn
sample are unioned into a single tag set.
We then perform exact-match and semantic deduplication at the tag level,
followed by greedy diversity sampling over tagged samples.
This procedure gives macro-level control over the balance between different
categories instead of deduplicating the raw pool as a whole.
On large components such as reasoning and general QA, only about 10--30\%
of the raw candidates are retained (the remaining 70--90\% are discarded),
whereas for smaller, less repetitive sources we keep 80--90\% of samples.
For each retained sample, the final assistant turn is regenerated with a
stronger teacher, \texttt{Qwen3-235B}, which improves answer quality and
stylistic consistency while keeping the original user input and context intact.

\paragraph{Ablations on datamix design}

We compare two variants of the midtraining corpus, both trained for
40B tokens with identical optimization settings
(Table~\ref{tab:middle_training_hparams}):

\begin{itemize}
    \item \textbf{Pre-train + instruct}: mixture including generic pre-training-style data
          (Common Crawl, Wikipedia, code) alongside instruction-formatted examples.
    \item \textbf{Instruct-only}: the same instruction pool but without additional raw
          pre-training sources, i.e., all examples follow an explicit
          instruction--response schema.
\end{itemize}

The instruct-only variant thus allocates more of the 40B-token budget to
high-quality instruction data, whereas the mixed variant spends a fraction
of tokens on generic web/code continuation.
We evaluate both models on a suite of Russian and multilingual
math/reasoning benchmarks, including ruAIME'24/25, ruMATH500, ruGPQA,
ruLCB, T-Math, and Arena-style
pairwise comparisons.
All evaluations are zero-shot; AIME-style metrics are computed as
$\text{avg@8}$ over 30 problems, and other benchmarks are run once due
to computational cost.

Table~\ref{tab:datamix-ablation} shows a representative subset of metrics.
Across most math and reasoning benchmarks, the instruct-only datamix
outperforms or matches the mixed variant despite using the same token
budget, and even early checkpoints from the instruct-only run are ahead
of the mixed model.
This is consistent with recent evidence that heavily pre-trained models
are harder to adapt via continual pre-training~\cite{springer2025overtrained},
especially when the additional data distribution differs from downstream tasks.

\begin{table}[!htbp]
    \centering
    \footnotesize
    \begin{tabular}{lcc}
        \toprule
        \textbf{Benchmark} &
        \textbf{PT+I} &
        \textbf{I-only} \\
        \midrule
        ruAIME'24                 & 0.60 & \textbf{0.67} \\
        ruAIME'25                 & 0.47 & \textbf{0.63} \\
        ruMATH500                 & 0.93 & \textbf{0.94} \\
        ruGPQA                    & 0.58 & \textbf{0.66} \\
        ruLCB                     & 0.53 & \textbf{0.55} \\
        T-Math                    & 0.49 & \textbf{0.50} \\
        Arena hard (think)        & 43.7 & \textbf{44.5} \\
        Arena wildchat ru (think) & 55.0 & \textbf{55.1} \\
        \bottomrule
    \end{tabular}
    \caption{Ablation on midtraining datamix design (zero-shot).
    ``PT+I'' denotes the pre-training+instruct mixture; ``I-only'' uses
    only instruction-formatted data.}
    \label{tab:datamix-ablation}
\end{table}

We did not run additional ablations such as training on original (non-regenerated)
answers or disabling InsTag deduplication.
In the first case, instruction data come from heterogeneous sources with
uneven answer quality and formats, and we found it undesirable to train
on such out-of-distribution completions.
In the second case, skipping deduplication would require regenerating answers
for a much larger pool of raw samples, significantly increasing computational
cost; we leave this exploration for future work.

\paragraph{Tokenizer adaptation and MERA results}

A key objective of midtraining is to adapt the model to a new, denser
tokenizer for Russian without degrading downstream quality.
To quantify the impact of the tokenizer choice, we train two 8B models on
the same midtraining datamix with identical optimization hyperparameters,
differing only in the tokenizer (original Qwen3 vs.\ T-pro 2.0).

Table~\ref{tab:mera-tokenizer-ablation} reports MERA scores for
these two variants.
The T-pro 2.0 tokenizer attains a macro-average score comparable to the
original one (0.574 vs.\ 0.560), with only small per-task differences in
both directions.
In other words, replacing the tokenizer with a denser Cyrillic
segmentation does not degrade general Russian-language performance on MERA,
which is the primary design goal of midtraining.

\begin{table}[!htbp]
    \centering
    \small
     \resizebox{\columnwidth}{!}{
    \begin{tabular}{lcc}
        \toprule
        \textbf{Task} & \textbf{Qwen} & \textbf{T-pro 2.0} \\
         & \textbf{tokenizer} & \textbf{tokenizer} \\
        \midrule
        USE          & \textbf{0.198}      & 0.191 \\
        MaMuRaMu     & 0.784              & \textbf{0.796} \\
        ruWorldTree  & 0.966/0.966        & 0.966/0.966 \\
        ruCodeEval   & 0.173/0.45/0.585   & \textbf{0.454/0.689/0.756} \\
        RCB          & 0.557/0.479        & \textbf{0.564/0.47} \\
        MathLogicQA  & 0.710              & \textbf{0.731} \\
        ruOpenBookQA & 0.897/0.897        & \textbf{0.922/0.923} \\
        RWSD         & \textbf{0.446}     & 0.250 \\
        CheGeKa      & 0.30/0.368         & \textbf{0.31/0.384} \\
        LCS          & \textbf{0.102}     & 0.096 \\
        PARUS        & 0.868              & \textbf{0.912} \\
        MultiQ       & \textbf{0.381/0.517} & 0.344/0.478 \\
        ruMultiAr    & \textbf{0.402}     & 0.400 \\
        ruTiE        & 0.788              & \textbf{0.798} \\
        ruModAr      & 0.515              & \textbf{0.627} \\
        \midrule
        AVG          & 0.560              & \textbf{0.574} \\
        \bottomrule
    \end{tabular}
    }
    \caption{MERA scores for an 8B model with the original Qwen3
    tokenizer and the Cyrillic-dense T-pro 2.0 tokenizer. Bold marks the better
    value per row.}
    \label{tab:mera-tokenizer-ablation}
\end{table}

The midtraining loss curve in Figure~\ref{fig:middle-training-loss}
further illustrates the adaptation process.
Training loss decreases sharply over the first $\approx 1$k steps
($\approx 4$B tokens) before plateauing, indicating that a substantial token
budget at a relatively high learning rate is required to adapt the model to
the new tokenizer beyond what the smaller, lower-LR SFT budget alone could
provide.

\section{T-Wix SFT dataset}\label{app_twix_sft}
\subsection{General part of T-Wix}\label{gen_sft}
The general part of the dataset consists of 468k diverse prompts collected from open-source data and high-quality translations of English-language datasets, subsequently deduplicated. The dataset is assembled to enhance the model’s capabilities in coding, mathematics, dialogue, and other competencies expected from a modern LLM.

\subsubsection{Data Preparation}

First and foremost, a corpus of 14M instructions (mostly in English) is compiled from various open-source datasets. To select the most useful samples, a data filtering pipeline is developed. It consists of several consecutive stages aimed at deduplication and ensuring high thematic, qualitative, and complexity diversity of the SFT dataset. We also perform
deduplication against benchmark datasets to ensure
that no benchmark examples leak into the training
corpus.

\paragraph{LSH and Embedding-Based Deduplication.}
At the initial stage, simple deduplication is performed using locality-sensitive hashing (LSH) and embedding-based similarity search to eliminate duplicated samples originating from different open-source datasets.

\paragraph{Thematic Tag Filtering.}
To ensure thematic balance, the \#\textsc{InsTag}-based filtering approach~\cite{lu2024instag} is employed. The pipeline uses a trained tagging model to extract thematic tags from each instruction.


For the present work, the tagger is trained using the Qwen2.5-7B~\cite{qwen2025qwen25technicalreport} model on multilingual data with a context length of up to 32k tokens, allowing tagging in both Russian and English, including long-context data. This modification substantially reduces translation overhead, as tagging and filtering can be applied directly to multilingual raw data without prior translation into English.


To improve thematic balance, an additional domain-balancing stage—“Domain \& Complexity Balancing”—is included, as the tagger primarily produces low-level thematic annotations.

\paragraph{Domain \& Complexity Balancing.}
In addition to fine-grained thematic filtering, a higher-level balance is introduced across major knowledge domains and difficulty levels within each domain. To achieve this, six domains — \textit{Math, Code \& Programming, Science, General Instruct, General Knowledge, Writing} — and three complexity levels — \textit{School, Student, Professor} — are defined.

Using large-scale LLM-assisted annotation, approximately 14M samples are automatically labeled with both domain and complexity tags. Subsequently, the dataset is balanced across domains and further normalized by difficulty within each domain to regulate the model’s output capabilities and ensure a uniform skill distribution.

This stage enables finer control over the resulting model’s generalization behavior, preventing overrepresentation of specific topics or difficulty levels.

\paragraph{Reward Model Filtering.}
In the subsequent stage, samples are filtered according to prompt quality using scores from the Reward Model (RM) described in \ref{rm_training}. For each of the datasets comprising the 14M instructions, an RM score is computed, and the bottom 10\% of samples with the lowest scores within each dataset are filtered out.

This step effectively removes “noisy” or low-quality samples that could negatively impact downstream model performance, preserving only high-quality and instructionally meaningful examples.

\paragraph{Instruction Following Difficulty Filtering.}
A further filtering stage based on Instruction Following Difficulty (IFD) scores is incorporated, following the approach introduced in \cite{ifd}. These scores quantify the difficulty a language model faces in following a given instruction. For the present work, IFD scores are computed relative to a midtraining checkpoint to reflect the model’s actual instruction-following capability. Samples with excessively high IFD values (>1.0) are discarded as overly complex or ambiguous, while those with very low IFD scores (<0.7) are filtered out as trivially simple.

This selective filtering makes it possible to retain the most challenging and instructionally rich examples—those that contribute most to improving the model’s instruction-following ability—while removing both overly simple and excessively difficult samples.

\paragraph{Multilingual Filtering and Translation.}
The multilingual setup enables filtering to be conducted directly on mixed-language raw data, reducing both the cost and time associated with preliminary translation. Only the final curated dataset is translated into Russian to ensure cross-lingual consistency.

\paragraph{Rejection Sampling and Generation.}
High dataset quality is further ensured through the use of top LLMs and rejection sampling. Each final training completion is produced using DeepSeek-V3 and Qwen-235B-A22B models, generating 8 candidate responses per instruction. These candidates are then filtered using RM scores to select the highest-quality outputs.

This approach not only eliminates translation artifacts present in the raw multilingual data but also results in substantially higher-quality responses compared to the original samples, thereby improving the overall consistency and instructional value of the dataset.

Overall, the combined multistage filtering pipeline ensures that the final SFT dataset is diverse, balanced, and composed of high-quality, instructionally valuable samples, free from data leakage and redundancy.

This approach allows the training process to remain balanced across domains (e.g., code and math) without bias toward any particular category.

\subsection{Long Context}

To enhance the model’s ability to process extended inputs, a dedicated long-context dataset is constructed. 

A diverse collection of long texts is selected from publicly available data for the pre-training phase, covering various domains such as \textit{education, technology, business, scientific literature}, and \textit{fiction}. The dataset is distributed across multiple context lengths ranging from 8k to 32k tokens, enabling the model to learn robustly across different input sizes. Using DeepSeek-V3 and Qwen-235B-A22B, a variety of prompts and responses are generated for each text, encompassing summarization, open- and closed-domain QA tasks, as well as reasoning-oriented datasets.

The resulting long-context dataset increment constitutes about 1\% of the total SFT training data in samples and 7.7\% in tokens, providing valuable coverage for instruction tuning under extended context conditions.

\subsection{Parallel Corpora}
To maintain strong English proficiency, parallel corpora are added to the SFT dataset — that is, instructional samples presented in English alongside their Russian counterparts. 

A series of experiments on the language share within the dataset shows that an optimal ratio is approximately 10\% English data relative to the total SFT mix.

\subsection{Reasoning part of T-Wix} \label{reas_sft}
To enhance reasoning capabilities in Large Language Models (LLMs) for the Russian language, a high-quality, reasoning-focused dataset is constructed through a targeted distillation pipeline. Rather than maximizing data volume, the pipeline prioritizes instructional value and appropriate task difficulty, ensuring that each retained sample provides meaningful learning potential for the target student model. The process starts from a broad collection of English-language reasoning instructions, which are subsequently translated into Russian, deduplicated, and carefully balanced across domains to support diverse and robust reasoning behaviors.

\paragraph{Initial Pool Generation and Deduplication.}
The initial pool of data is constructed from approximately 450k high-quality English-language reasoning instructions, drawn from established open-source datasets (e.g., Open-R1 \cite{openr1}, Nvidia/AceReason-Math \cite{chen2025acereason}, Nvidia/Nemotron \cite{bercovich2025llamanemotronefficientreasoningmodels}), covering general knowledge, mathematics, natural sciences, and code generation.

Domain Distribution:
\begin{itemize}
    \item 60\% general knowledge and open-ended reasoning (to establish fluent, structured reasoning in Russian),
    \item 10\% verifiable mathematics (e.g., arithmetic, algebra),
    \item 10\% open-ended mathematics (e.g., proofs, conceptual explanations),
    \item 15\% natural sciences (physics, chemistry, biology),
    \item 5\% code-related reasoning.
\end{itemize}

After the initial collection, these English-language instructions are translated into Russian. As in the general part of T-Wix, deduplication is applied to eliminate near-duplicates and ensure sample uniqueness.

\paragraph{Reward-Based Completion Evaluation.}
To mitigate the stochasticity inherent in LLM generation \cite{wang2025effectsamplingdiversityscaling, atil2025nondeterminismdeterministicllmsettings}, 8 diverse completions are generated per instruction by both:
\begin{itemize}
    \item The teacher model (Qwen3-235B-A22B),
    \item The student model (midtraining checkpoint).
\end{itemize}
This yields 16 completions per instruction, which are independently scored by a trained reward model (RM). The inclusion of student generations enables direct assessment of the model’s current reasoning capability on each task, while the teacher generations provide high-quality reference behaviors. This multi-generation approach provides a more robust statistical picture of the model’s performance on each instruction.

\paragraph{Statistical Filtering Based on Reward Stability.}
Instructions exhibiting high variance in RM scores across generations are discarded, as they reflect ambiguous or unstable evaluation signals. Additionally, instructions for which the student model consistently receives very low RM scores—even with low variance—are excluded, as they lie beyond the student’s current learning capacity and are unlikely to support effective knowledge transfer~\cite{xiong2025minimalistapproachllmreasoning, liu2025itrickstrapsdeep}.
\paragraph{Mean Reward-Based Selection Within the Zone of Proximal Development.}
To operationalize the pedagogical principle of the zone of proximal development (ZPD)~\cite{cui2025investigatingzoneproximaldevelopment}, the RM scores for the 8 teacher and 8 student completions per instruction are aggregated by computing their respective means. The average reward of the teacher responses and the average reward of the student responses are then used to estimate the reasoning gap.

Samples are selected based on the absolute difference between these mean rewards. A small difference indicates that the student already performs comparably to the teacher (suggesting limited learning potential), whereas a very large difference implies that the task lies beyond the student’s capabilities (making distillation ineffective). Only samples with a moderate gap in mean RM scores—neither too small nor too large—are retained. This ensures that the selected samples are challenging enough to drive improvement, yet sufficiently within reach for successful knowledge transfer.

\paragraph{Final Completion Selection.}
For the final training targets, teacher-generated completions are selected as follows:
\begin{itemize}
    \item For \textit{verifiable} instructions (e.g., mathematical problems), factually incorrect completions are first filtered out; among the remaining correct ones, the completion with the highest RM score is chosen.
    \item For \textit{open-ended} instructions, the shortest reasoning trace among the top-3 RM-ranked teacher completions is selected. This encourages the student to learn concise and non-redundant reasoning patterns~\cite{sui2025stopoverthinkingsurveyefficient}.
\end{itemize}

This pipeline yields a high-quality reasoning dataset of approximately 30k samples, consisting of 90\% Russian and 10\% English instructions, consistent with the overall language strategy of T-Wix. By design, the dataset emphasizes stable, diverse, and pedagogically optimal reasoning traces in Russian across multiple domains, effectively balancing task difficulty with learning potential.

\subsection{T-Wix dataset analytics}

\begin{figure}[!htbp]
    \centering
    \includegraphics[width=0.85\linewidth]{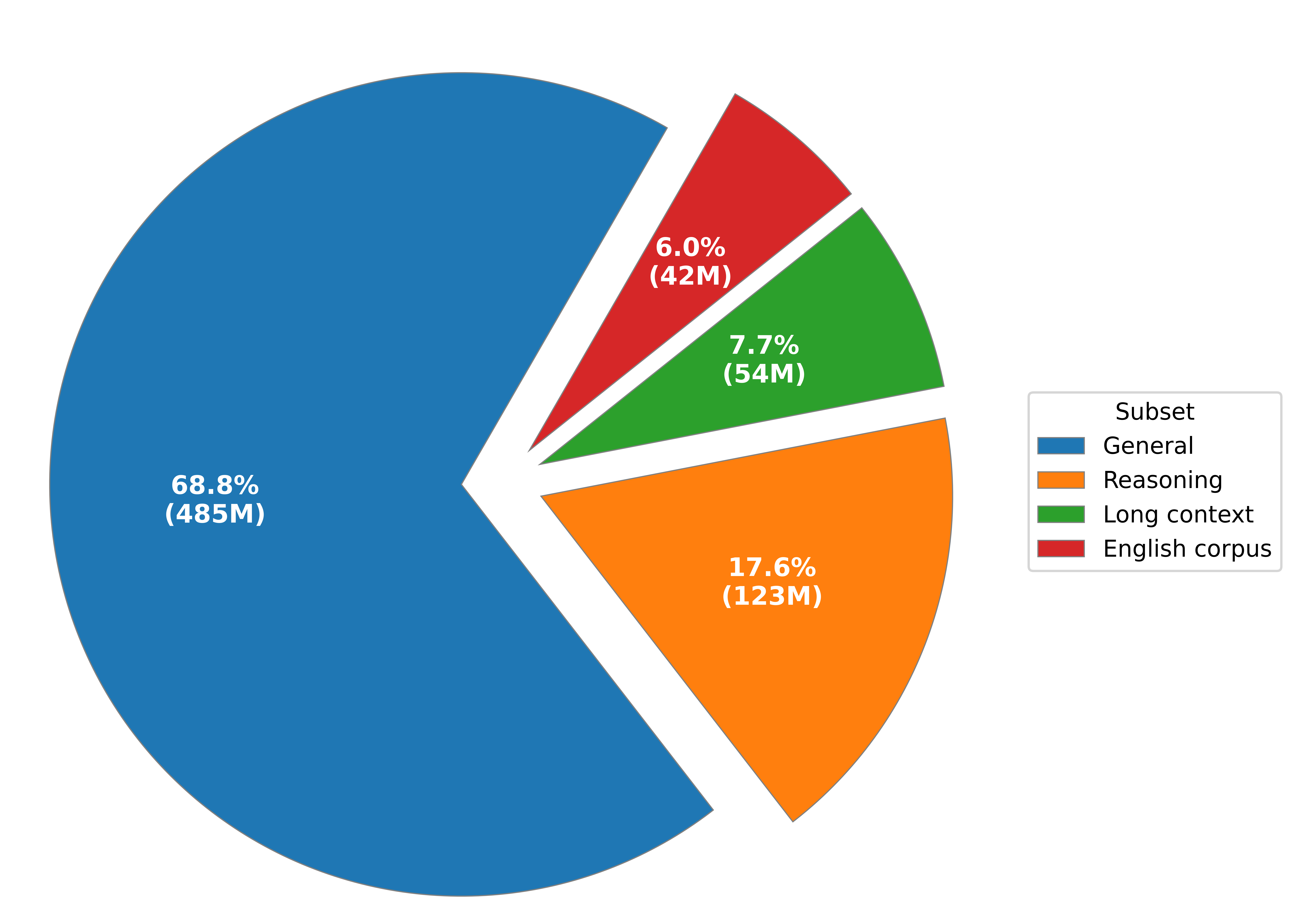}
    \caption{Token distribution in T-Wix. Token counts were computed using tiktoken with o200k base tokenizer.}
    \label{fig:domain_balance}
\end{figure}

The final distribution of data in the SFT dataset (T-Wix) are presented in Figure~\ref{fig:domain_balance}. The total size is 500k samples.

\subsection{SFT Training Recipe}

The SFT stage took 9 hours on 4 nodes with 8×H100 GPUs for T-pro 2.0, using gradient checkpointing and FSDP, as well as packing samples into 32k-token contexts without truncation. After a series of experiments, the optimal fine-tuning hyperparameters were selected, as described in Table~\ref{tab:sft-hparams}.

\begin{table}[!htbp]
    \small
    \centering
    \resizebox{\columnwidth}{!}{
    \begin{tabular}{l r}
        \toprule
        \textbf{Hyperparameter} & \textbf{Value} \\
        \midrule
        Global batch size (samples) & 32 \\
        Max context length & 32k \\
        Number of training epochs & 2 \\
        \midrule
        Optimizer & Adam \cite{kingma2017adammethodstochasticoptimization} \\
        Adam betas & $(0.9, 0.95)$ \\
        Adam $\epsilon$ & $10^{-12}$ \\
        Learning rate & 1e-6 \\
        Learning rate scheduler & cosine \\
        Warmup ratio & 0.1 \\
        Gradient clipping & max norm 2.0 \\
        \midrule
        Precision & BF16 \\
        \bottomrule
    \end{tabular}
    }
    \caption{Hyperparameters used for SFT training.}
    \label{tab:sft-hparams}
\end{table}

\section{Preference tuning} \label{pref_tuning}
To enhance alignment beyond supervised fine-tuning, an on-policy Direct Preference Optimization (DPO) procedure is applied. Recent work shows that on-policy preference optimization offers more stable and reliable alignment gains than off-policy alternatives, as it learns directly from the model’s own generative distribution and therefore avoids distribution shift while targeting realistic error modes \cite{dpo1, dpo2}.

For each instruction, the SFT-trained model produces 16 completions. All candidates are scored using the RM described in App.~\ref{rm_training}, and preference pairs are constructed by selecting the highest- and lowest-scoring completions. This contrastive selection yields stable and informative training pairs by removing low-signal, ambiguous comparisons.

The DPO dataset is constructed from filtered SFT data (T-Wix). A total of 100k preference pairs is formed, consisting of:
\begin{itemize}
    \item 90k sampled from the General SFT part,
    \item 10k sampled from the Reasoning SFT part.
\end{itemize}

In addition, cross-subset augmentation is applied to enrich preference diversity: 4k samples from the General subset are paired with reasoning-style reformulations, while 6k samples from the Reasoning subset are converted into general-style instructions. This yields a smoother distribution of reasoning complexity without altering the intended emphasis of each subset.

The resulting on-policy DPO stage improves the model’s alignment, coherence, and reasoning structure while preserving broad general-purpose capabilities.

\subsection{Preference Training Recipe}
\label{dpo_receipe}

The DPO stage required 28 hours of training on 4 nodes with 8×H100 GPUs. The training was carried out using sequence parallelism, which enabled efficient distribution of computation across devices. The hyperparameters listed in Table~\ref{tab:preference-hparams} were identified as optimal.

\begin{table}[!htbp]
    \small
    \centering
    \begin{tabular}{l r}
        \toprule
        \textbf{Hyperparameter} & \textbf{Value} \\
        \midrule
        Global batch size (samples) & 128 \\
        Max context length & 32k \\
        Number of training epochs & 1 \\
        \midrule
        Optimizer & AdamW  \\
        Adam betas & $(0.9, 0.95)$ \\
        Adam $\epsilon$ & $10^{-12}$ \\
        Weight decay & 0.01 \\
        Learning rate & 1e-7 \\
        Learning rate scheduler & cosine \\
        Warmup ratio & 0.05 \\
        Gradient clipping & max norm 2.0 \\
        \midrule
        Precision & BF16 \\
        Loss type & DPO \\
        DPO beta & 0.5 \\
        \bottomrule
    \end{tabular}
    \caption{Hyperparameters used for preference tuning}
    \label{tab:preference-hparams}
\end{table}
\section{Reward Model} \label{rm_training}

\paragraph{Tournament-Based Synthetic Preference Data Generation}

To construct a high-quality reward model, it is essential to obtain reliable preference data~---pairs of model completions ranked according to their relative quality. Direct pairwise annotation across all available completions, however, is computationally expensive and inefficient. To address this, similar to the knockout-tournament method introduced by \citet{liu2025PairJudge}, we propose a tournament-based preference generation approach that substantially reduces the number of required comparisons while preserving the informativeness of the resulting preference signal.

Each tournament comprises $n$ participants, randomly sampled from the pool of available models. For each instruction, every model generates a completion, and the tournament bracket is constructed according to model category~---for instance, small-scale models (7B–13B) are paired against models of similar scale, and reasoning-oriented models compete within the same subclass. This grouping strategy ensures that comparisons are made between models of comparable generative quality, encouraging the reward model to learn fine-grained distinctions rather than relying on trivial cases where one output is clearly superior (e.g., when a large model is compared with a small size model, a comparison might yield an obvious outcome~---the larger model would consistently produce more coherent and contextually appropriate responses, leaving little room for the reward model to learn subtle differences).

Each round of the tournament consists of a single instruction and the corresponding completions generated by the competing models. An external LLM, not participating in the tournament, is employed as judge to determine the preferred completion for each matchup. To avoid positional bias, each pair of completions is evaluated in both possible orders, and samples exhibiting positional bias are excluded from the final training set.

At the completion of each single-elimination tournament with $n$ participants, a total of $\frac{n}{2} \log_2 N$ preference pairs are obtained. This result comes from the hierarchical structure of the tournament: in each round, half of the remaining participants compete, producing $\frac{n}{2}$ new pairwise outcomes (both direct and transitive). Because a tournament with $n$ participants requires $\log_2 N$ rounds to determine a winner, the total number of inferred preference pairs accumulates to $\frac{n}{2} \log_2 N$.

Each round contributes the same number of new known preferences because every winner’s new victory also establishes transitive relationships over all opponents defeated in earlier rounds. For instance, if player A beats player B in the final, it is implied that A outperform every player that B previously defeated. Consequently, even though only $n - 1$ matches are directly played, the tree-like transitive structure allows many additional indirect comparisons to be inferred.

This process produces preference set dense enough to capture comparative information among many participants, yet far more efficient than exhaustively comparing every possible pair (which would require $\frac{n(n-1)}{2}$ comparisons). This tournament-based approach yields an informative preference dataset while significantly reducing annotation complexity.

\paragraph{Reward Model Training}

The reward model is based on Qwen3-32B~\cite{yang2025qwen3technicalreport} with a regression head to produce a single preference score for each completion. Training follows the Bradley–Terry \cite{10.1093/biomet/39.3-4.324} formulation, which models the probability of one completion being preferred over another as a logistic function of their respective scores. All training is conducted with a maximum sequence length of 32k tokens, leveraging Ulysses sequence parallelism \cite{10.1145/3662158.3662806} to efficiently support long-context optimization. Data preprocessing, batching, and distributed training are managed through the TurboAlignment library \cite{turboalignment}.

\paragraph{Evaluation}

For intrinsic evaluation, we adapt RewardBench~2~\cite{malik2025rewardbench2advancingreward} to Russian by translating the original benchmark and report standard leaderboard metrics. For downstream evaluation, we additionally construct a Best-of-$N$ selection benchmark on top of the Arena-Hard-RU instruction set to assess the reward model under realistic generation scenarios. In this setting, the base model produces $N$ candidate completions per instruction, and the reward model selects the highest-scoring (best@N) and lowest-scoring (worst@N) outputs. These selections are then evaluated using Arena-Hard, allowing us to measure the alignment between reward-model rankings and externally validated quality. We further report the $\Delta_{\text{BoN}}$ metric (best@N -- worst@N) to quantify discriminative capacity. Although our model performs comparably to existing open-source reward models on the translated RewardBench~2, it demonstrates a clear advantage on our Best-of-$N$ Arena-Hard benchmark. As shown in Table~\ref{tab:bon-comparison}, our model obtains the highest $\Delta_{\text{BoN}}$ score, reflecting the strongest separation between high- and low-quality completions.

\begin{table}[!htbp]
\centering
\small
\resizebox{\columnwidth}{!}{%
\begin{tabular}{@{}lccc@{}}
\toprule
RM-model & best@8$\uparrow$ & worst@8\(\downarrow\) & $\Delta_{\text{BoN}}$ \\ 
\midrule
\textbf{Qwen3-32B-RM (Ours)} & \textbf{92.69} (-0.99) & \textbf{70.48} (+2.34) & \textbf{22.21} \\
Llama-3.3-Nemotron-70B-Reward-Multilingual\textsuperscript{1} & 85.93 (-1.93) & 84.91 (+1.85) & 1.02 \\
Skywork-Reward-Gemma-2-27B\textsuperscript{2} & 89.05 (-1.6)  & 74.35 (+2.07) & 14.70 \\
Skywork-Reward-V2-Llama-3.1-8B\textsuperscript{3} & 90.49 (-1.43) & 77.31 (+1.77) & 13.18 \\
Llama-3.1-Tulu-3-70B-SFT-RM-RB2\textsuperscript{4} & 87.37 (-1.86) & 78.47 (+1.76) & 8.90 \\
\bottomrule
\end{tabular}
}
\caption{
Best-of-$N$ ($N=8$) evaluation on Arena-Hard-RU. 
We report win rates for the highest- (best@8) and lowest-scoring (worst@8) completions selected by each reward model, and their difference $\Delta_{\text{BoN}} = \text{best@8} - \text{worst@8}$, which measures discriminative capacity.
\textsuperscript{1}\citet{wang2025helpsteer3preferenceopenhumanannotatedpreference}, \textsuperscript{2}\citet{liu2024skywork}, \textsuperscript{3}\citet{liu2025skywork}, \textsuperscript{4}\citet{malik2025rewardbench2advancingreward}}
\label{tab:bon-comparison}
\end{table}

\begin{table*}[!htbp]
\centering
\small
\resizebox{\textwidth}{!}{%
\begin{tabular}{@{}llccccc|c@{}}
\toprule
\multirow{2}{*}{Category} & \multirow{2}{*}{Model} & \multicolumn{6}{c}{RewardBench 2 (RU)} \\ 
\cmidrule(l){3-8}
               &  & Fact. & Focus & Math  & Prec. IF & Safety & Total \\ 
\midrule
\multirow{3}{*}{\shortstack[l]{Comparison with\\Existing RMs}} 
& \cellcolor{blue!10}\textbf{Qwen3-32B-RM (Ours)} & \cellcolor{blue!10}0.66 & \cellcolor{blue!10}0.87 & \cellcolor{blue!10}0.62 & \cellcolor{blue!10}0.42 & \cellcolor{blue!10}0.89 & \cellcolor{blue!10}0.69 \\
& Skywork-Reward-V2-Llama-3.1-8B & 0.68 & \textbf{0.88} & 0.65 & \textbf{0.45} & 0.79 & 0.69 \\
& Skywork-Reward-Gemma-2-27B & 0.69 & \textbf{0.88} & 0.64 & 0.40 & \textbf{0.92} & \textbf{0.7} \\ 
& Llama-3.1-Tulu-3-70B-SFT-RM-RB2  & 0.72	& 0.74	& \textbf{0.69}	& 0.41	& 0.76 & 0.66 \\
& Llama-3.3-Nemotron-70B-Reward-Multilingual  & \textbf{0.73}	& 0.85	& 0.62	& 0.41	& 0.86 & 0.69 \\
\midrule
\multirow{2}{*}{\shortstack[l]{Judge Model\\Ablation}} 
& Qwen-3-RM-8B-DeepSeek-V3   & \textbf{0.478} & \textbf{0.756} & 0.598 & 0.341 & 0.736 & \textbf{0.581} \\
& Qwen-3-RM-8B-Qwen3-235B-A22B & 0.467 & 0.324 & \textbf{0.688} & \textbf{0.350} & \textbf{0.840} & 0.533 \\
\midrule
\multirow{2}{*}{\shortstack[l]{Transitive Samples\\Ablation}} 
& Qwen3-8B-RM w/o transitive & 0.467 & 0.324 & 0.688 & 0.350 & 0.840 & 0.533 \\
& Qwen3-8B-RM w/ transitive & \textbf{0.505} & \textbf{0.453} & \textbf{0.704} & \textbf{0.413} & \textbf{0.860} & \textbf{0.587} \\
\bottomrule
\end{tabular}
}
\caption{Evaluation results on RewardBench 2 (RU). We compare our Qwen3-32B-RM against existing reward models (top), analyze the impact of different judge models for preference annotation (middle), and study the effect of tournament-derived transitive preference samples during training (bottom). Bold indicates best performance within each category.}
\label{tab:eval-results-rewardbench2-combined}
\end{table*}

\paragraph{Prompt selection}
Furthermore, in the process of synthetic data generation, we evaluated a range of prompting strategies derived from the JudgeBench~\cite{tan2025judgebench}. Empirical analysis indicates that the Google Vertex prompt yields superior evaluation quality in different benchmarks (see Table~\ref{tab:prompt-rewardbench2}), particularly on RewardBench~2 (RU). This improvement underscores the sensitivity of LLM-based evaluators to prompt design and highlights the importance of selecting domain-appropriate judging configurations for reliable preference data generation.

\begin{table}[!htbp]
\centering
\small
\resizebox{\columnwidth}{!}{\begin{tabular}{@{}lcccccc@{}}
\toprule
\multirow{2}{*}{Prompt} & \multicolumn{6}{c}{RewardBench 2 (RU)} \\ 
\cmidrule(l){2-7}
               & Fact. & Focus & Math  & Prec. IF & Safety & Total \\ 
\midrule
Skywork        & 0.636 & 0.782 & \textbf{0.834} & 0.394 & 0.881 & 0.706 \\
Arena Hard     & 0.653 & 0.638 & 0.762 & 0.349 & 0.899 & 0.660 \\
Google Vertex  & \textbf{0.741} & \textbf{0.846} & 0.830 & \textbf{0.549} & \textbf{0.915} & \textbf{0.776} \\
Prometheus 2   & 0.600 & 0.622 & 0.743 & 0.432 & 0.790 & 0.637 \\
Chat-Eval      & 0.667 & 0.781 & 0.795 & 0.478 & 0.831 & 0.710 \\
\bottomrule
\end{tabular}
}
\caption{Assessing the role of prompt selection in RewardBench~2 (RU).}
\label{tab:prompt-rewardbench2}
\end{table}

\subsection{Reward Model Analysis}
\label{rm_analysis}

Our experiments reveal that DeepSeek-V3~\cite{deepseekai2025deepseekv3technicalreport} demonstrates superior judgment capabilities in open-domain and conversational (chat) tasks, whereas Qwen3-235B-A22B exhibits stronger performance in mathematical, code-related and other domains (see Table~\ref{tab:eval-results-rewardbench2-combined}).

\paragraph{Ablation on Transitive Samples.}
An ablation study was conducted to evaluate the contribution of transitive preference pairs. Removing transitive samples led to a consistent degradation across all evaluation metrics (see Table~\ref{tab:eval-results-rewardbench2-combined}), suggesting that inferred pairwise relationships enrich the preference signal and improve the model’s generalization to unseen instructions. Conversely, adding additional transitive samples beyond the first closure continued to yield marginal but positive improvements.

\paragraph{Length Sensitivity and Distribution Effects.}
A further observation concerns the length distribution between chosen and rejected completions. RewardBench~2 (RU) exhibits a substantial drop in evaluation quality when the distribution becomes skewed~---specifically, when longer or shorter completions dominate. This imbalance appears to induce a length-based bias in the reward model, leading it to systematically favor responses of a particular size rather than quality.

For instance, Qwen3-235B-A22B as a judge displayed a pronounced length bias, consistently preferring longer completions regardless of their semantic quality. This highlights the importance of maintaining a balanced length distribution during preference data generation and tournament construction to prevent undesirable inductive shortcuts in the reward model.

\section{Speculative Decoding Implementation}
\label{app:speculative_decoding}

To mitigate the sequential latency of autoregressive generation, we integrate an EAGLE-based speculative decoding module~\citep{Li2024EAGLE} into T-pro 2.0. This setup employs a lightweight draft model to propose candidate tokens in parallel, which are subsequently verified by the target model to ensure the output distribution remains identical to standard decoding~\citep{Leviathan2023}.

\paragraph{Architecture and Objective.}
Our draft model utilizes a single decoder layer augmented with an FR-Spec component~\cite{zhao-etal-2025-fr}, based on the Llama 2 architecture and implemented via SGLang~\citep{Gu2024SGLang}. Unlike standard approaches that replicate the full target architecture, this model approximates essential hidden-state dynamics. The training objective combines a smoothed $L_1$ loss (MAE and MSE) for hidden state reconstruction with KL divergence to align the draft token distribution with the target model.

\paragraph{Data and Training Pipeline.}

We evaluated three data pipelines: offline labeling (I/O bound), chunked streaming (network bound), and online labeling. We adopted Online Labeling for the final setup. Although this increases HBM footprint by requiring the frozen target model to reside in memory, it yields the highest Tensor Core utilization.

Training was performed on a single node with 8$\times$H100 GPUs. The frozen verifier used Tensor Parallelism, while the EAGLE draft model utilized Distributed Data Parallelism. Full training hyperparameters are listed in Table~\ref{tab:eagle-hparams}.

\begin{table}[!htbp]
    \small
    \centering
    \begin{tabular}{l r}
        \toprule
        \textbf{Hyperparameter} & \textbf{Value} \\
        \midrule
        Hardware & 8$\times$H100 (80GB) \\
        Verifier parallelism & TP=2 \\
        Draft model parallelism & DDP (world size=8) \\
        \midrule
        Batch size & 32 \\
        Learning rate & 3e-5 \\
        Number of epochs & 4 \\
        Learning rate scheduler & cosine \\
        Warmup steps & 100 \\
        Weight decay & 0.01 \\
        Optimizer & AdamW \\
        Data type & BF16 \\
        TF32 & enabled \\
        \bottomrule
    \end{tabular}
    \caption{Hyperparameters used for EAGLE draft model training.}
    \label{tab:eagle-hparams}
\end{table}

\paragraph{Deployment and Results.}
Deployed via SGLang using EAGLE-2's dynamic draft tree~\citep{li2024eagle2}, the system achieves significant latency reductions. Table~\ref{tab:benchmark_results} highlights speedups up to $2.28\times$ on reasoning tasks (T-Math) and consistent gains on ruMT-Bench. Table~\ref{tab:mmlu_domains} illustrates domain-specific performance on  ruMMLU-
Pro, where Math and Engineering domains show the highest acceptance lengths ($\sim$3.7) and speedups ($\sim$2.0$\times$). Future work will focus on draft model quantization and integrating EAGLE 3~\citep{li2025eagle3}.

\begin{table}[!htbp]
\centering
\small
\resizebox{\columnwidth}{!}{%
\begin{tabular}{lcccc}
\toprule
\textbf{Benchmark} & \textbf{Temp.} & \textbf{Mode} & \textbf{Speedup} & \textbf{\shortstack{ Acceptance\\Length}} \\
\midrule
\multirow{4}{*}{ruMT-Bench} 
 & 0 & No Think & 2.05 & 3.55 \\
 & 0 & Think & 1.86 & 3.37 \\
 & 0.8 & No Think & 1.79 & 3.31 \\
 & 0.8 & Think & 1.69 & 3.10 \\
\midrule
\multirow{4}{*}{ruAlpaca} 
 & 0 & No Think & 1.78 & 3.23 \\
 & 0 & Think & 1.77 & 3.20 \\
 & 0.8 & No Think & 1.61 & 2.94 \\
 & 0.8 & Think & 1.57 & 2.85 \\
\midrule
\multirow{4}{*}{ruCodeEval} 
 & 0 & No Think & 2.26 & 4.09 \\
 & 0 & Think & 2.07 & 3.76 \\
 & 0.8 & No Think & 2.15 & 3.93 \\
 & 0.8 & Think & 1.84 & 3.34 \\
\midrule
\multirow{2}{*}{T-Math} 
 & 0 & Think & 2.28 & 4.14 \\
 & 0.8 & Think & 2.25 & 4.01 \\
\bottomrule
\end{tabular}
}
\caption{Performance metrics for T-pro-2.0-eagle across different benchmarks, temperatures, and reasoning modes. Comparison of Speedup and Acceptance Length with and without Eagle.}
\label{tab:benchmark_results}
\end{table}

\begin{table}[!htbp]
\centering
\resizebox{\columnwidth}{!}{%
\begin{tabular}{lcccc}
\toprule
\multirow{2}{*}{\textbf{Domain}} & \multirow{2}{*}{\textbf{Speedup}} & \multirow{2}{*}{\textbf{\shortstack{Accept.\\Length}}} & \multicolumn{2}{c}{\textbf{TPS}} \\
\cmidrule(lr){4-5}
 & & & \textbf{w/o Eagle} & \textbf{w/ Eagle} \\
\midrule
Biology & 1.68 & 3.00 & 108.22 & 181.86 \\
Business & 2.00 & 3.63 & 107.83 & 216.49 \\
Computer Sci. & 1.89 & 3.37 & 107.99 & 204.22 \\
Economics & 1.72 & 3.07 & 108.26 & 185.80 \\
Engineering & 2.00 & 3.60 & 106.96 & 214.37 \\
Health & 1.67 & 2.98 & 108.29 & 181.00 \\
History & 1.52 & 2.72 & 108.15 & 164.17 \\
Law & 1.51 & 2.69 & 108.03 & 163.17 \\
Math & 2.06 & 3.70 & 107.88 & 221.96 \\
Philosophy & 1.62 & 2.88 & 108.37 & 175.29 \\
Physics & 1.96 & 3.50 & 107.60 & 210.60 \\
Psychology & 1.65 & 2.85 & 108.38 & 179.03 \\
Chemistry & 2.04 & 3.66 & 107.56 & 219.20 \\
\bottomrule
\end{tabular}%
}
\caption{Performance metrics for T-pro-2.0-eagle across ruMMLU-
Pro domains (Temperature 0.8, Thinking mode, Batch size=1).}
\label{tab:mmlu_domains}
\end{table}
\section{T-Math benchmark}
\label{app:tmath}

\begin{table*}[t]
    \centering
    \small
    \begin{tabular}{p{0.8cm} p{11cm} p{1.8cm}}
        \toprule
        \textbf{\#} & \textbf{Problem statement (translated from Russian for readability)\footnotemark} & \textbf{Answer} \\
        \midrule
        1 &
        \textbf{Combinatorics / logic.}
        In a tournament there are 20 players and 10 referees.
        After each game, the participants of that game take a photograph together
        with the referee.
        After the tournament it turned out that for some people it is impossible to
        determine whether they are a player or a referee (based only on the set of
        photos they appear in).
        What is the maximum possible number of such people? &
        $2$ \\
        \midrule
        2 &
        \textbf{Number theory / arithmetic constructions.}
        Using any number of coins of denominations 1, 2, 5 and 10 roubles, together
        with (free) parentheses and the four arithmetic operations, construct an
        expression whose value is $2009$, while spending as little money as possible.
        In the answer, write the minimum possible total value of the coins used
        (i.e., the minimum amount of money you need to ``spend''). &
        $23$ \\
        \midrule
        3 &
        \textbf{Geometry, olympiad level.}
        In triangle $ABC$ with side lengths $AB = 3$, $BC = 4$, $CA = 5$, we mark
        pairs of points on its sides:
        points $C_1$ and $C_2$ on side $AB$,
        points $A_1$ and $A_2$ on side $BC$,
        and points $B_1$ and $B_2$ on side $CA$.
        Inside triangle $ABC$ there is a point $P$ such that triangles
        $PA_1A_2$, $PB_1B_2$ and $PC_1C_2$ are congruent and equilateral.
        Find the area of the convex hexagon with vertices
        $A_1$, $A_2$, $B_1$, $B_2$, $C_1$, $C_2$.
        If necessary, round your answer to two decimal places. &
        $3.34$ \\
        \bottomrule
    \end{tabular}
    \caption{Example problems from the T-Math benchmark. Statements are translated
    from the original Russian for readability; see the dataset for the original wording
    and full benchmark specification.}
    \label{tab:tmath-examples}
\end{table*}

\begin{table*}[!htbp]
\centering
\small
\begin{tabular}{@{}lccccc@{}}
\toprule
\textbf{Model} & \textbf{AIME 2024} & \textbf{AIME 2025} & \textbf{MATH-500} & \textbf{GPQA Diamond} & \textbf{LCB} \\
\midrule
\textit{Open Source Models (27B-32B class)} &  &  &  &  &  \\
\cellcolor{blue!10}\textbf{T-pro 2.0 (Ours)} & \cellcolor{blue!10}\underline{0.765} & \cellcolor{blue!10}\underline{0.679} & \cellcolor{blue!10}\textbf{0.966} & \cellcolor{blue!10}\underline{0.641} & \cellcolor{blue!10}\underline{0.556} \\
Qwen3-32B & \textbf{0.808} & \textbf{0.725} & \underline{0.961} & \textbf{0.668} & 0.546 \\
RuadaptQwen3-32B-Instruct & 0.692 & 0.604 & 0.948 & 0.596 & 0.489 \\
Gemma 3 27B & 0.260 & 0.221 & 0.882 & 0.515 & 0.246 \\
DeepSeek-R1-Distill-Qwen-32B & 0.706 & 0.573 & 0.950 & 0.621 & \textbf{0.572} \\
\midrule
\multicolumn{6}{l}{\textit{Open Source Larger Scale \& Proprietary Models}} \\
DeepSeek-V3            & 0.52  & 0.285 & 0.942 & 0.655 & 0.405 \\
DeepSeek-R1            & \textbf{0.914} & \textbf{0.875} & \textbf{0.983} & \textbf{0.813} & \textbf{0.770} \\
YandexGPT5-Pro         & 0.117 & 0.090 & 0.776 & 0.434 & 0.272 \\
GigaChat 2 Max         & 0.110 & 0.058 & 0.742 & 0.449 & 0.272 \\
o4-mini (medium)       & \underline{0.800} & \underline{0.819} & \underline{0.974} & \underline{0.783} & \underline{0.757} \\
GPT-4o                 & 0.098 & 0.065 & 0.762 & 0.545 & 0.246 \\
    
\bottomrule
\end{tabular}
\caption{Comparison of models on English advanced reasoning benchmarks.}
\label{tab:en_reasoning2}
\end{table*}

T-Math\footnote{\url{https://huggingface.co/datasets/t-tech/T-math}} is a
Russian math reasoning benchmark constructed from high-school olympiad
problems.
It contains 331 tasks drawn from the All-Russian School Olympiad and the
Moscow Olympiad in mathematics over the period 1998–2025.
All items are single-answer problems with numeric gold solutions, which makes
the benchmark suitable for automatic evaluation of long-chain mathematical
reasoning.

Problem statements and ground-truth answers are extracted from PDF
collections using the Qwen2.5-VL-72B-Instruct \cite{bai2025qwen25vltechnicalreport}
model.
The raw pool is then filtered with an LLM-based checker to discard
(i) tasks requiring multiple answers, (ii) problems without a unique correct
answer, (iii) theorem-style questions where the main goal is to prove a
statement, (iv) tasks whose solutions are non-numeric, and
(v) items that cannot be solved without access to auxiliary figures.
Next, medium-difficulty tasks on which \texttt{Qwen3-8B} achieves
near-perfect pass@16 are removed to focus the benchmark on genuinely
challenging instances.
Finally, both the question texts and the verifiable answers are manually
reviewed against the original olympiad sources.
Evaluation uses a standardized answer format (final answer wrapped in
\verb|\boxed{}|) and the \texttt{math\_verify} library\footnote{\url{https://github.com/huggingface/Math-Verify}}
to compare predicted and reference expressions.

Table~\ref{tab:tmath-results} reports pass@1 scores for several strong
reasoning models.
Although frontier systems such as \texttt{o4-mini-high}, DeepSeek-R1 and
Gemini~2.5 Pro achieve competitive performance, the benchmark remains far
from saturated, with none of the models exceeding 0.75 pass@1.
\begin{table}[!htbp]
    \centering
    \small
    \begin{tabular}{lc}
        \toprule
        \textbf{Model} & \textbf{pass@1} \\
        \midrule
        o4-mini-high        & \textbf{0.73} \\
        DeepSeek-R1-0528    & \underline{0.71} \\
        Gemini-2.5-Pro      & 0.70 \\
        Claude Sonnet 4     & 0.56 \\
        \cellcolor{blue!10}T-pro 2.0        & \cellcolor{blue!10}0.54 \\
        Qwen3-32B           & 0.53 \\
        \bottomrule
    \end{tabular}
    \caption{Pass@1 accuracy on the T-Math benchmark (331 problems).}
    \label{tab:tmath-results}
\end{table}
\section{Additional Evaluations} \label{en_results_benches}

As shown in Table~\ref{tab:en_reasoning2}, the model preserves strong English reasoning ability despite being primarily optimized for Russian. Within the 27B–32B class, it remains closely aligned with the Qwen3-32B baseline: on MATH-500 it slightly improves accuracy, and on AIME 2024/2025 and GPQA the margins stay narrow. Performance is also competitive with reasoning-distilled systems such as DeepSeek-R1-Distill-Qwen-32B, outperforming them on several metrics. Overall, these results indicate that the Cyrillic-focused tokenizer and our training pipeline do not meaningfully degrade English performance, maintaining robust cross-lingual generalization with minimal loss on advanced benchmarks.

\end{document}